\documentclass[journal]{IEEEtran}

\ifCLASSINFOpdf
\usepackage[pdftex]{graphicx}
\usepackage{multirow}
\usepackage{amsmath}
\usepackage{caption}
\usepackage{xcolor} 
\usepackage{color} 
\usepackage{verbatim} 
\newcommand{\tabincell}[2]{\begin{tabular}{@{}#1@{}}#2\end{tabular}}
\else
\fi

\begin{document}
	%
	\title{Deep Video Super-Resolution using HR Optical Flow Estimation}
	
	
	\author{\IEEEauthorblockN{Longguang Wang,
			Yulan Guo,
			Li Liu,
			Zaiping Lin, 
			Xinpu Deng, and
			Wei An}
		\thanks{
			Longguang Wang, Yulan Guo, Zaiping Lin, Xinpu Deng and Wei An are with the College of Electronic Science and Technology, National University of Defense Technology, Changsha, 410073, China (e-mail: wanglongguang15@nudt.edu.cn; yulan.guo@nudt.edu.cn;  linzaiping@nudt.edu.cn; dengxinpu@nudt.edu.cn; anwei@nudt.edu.cn). Yulan Guo is also with the School of Electronics and Communication Engineering, Sun Yat-sen University, Guangzhou, 510275, China.
			Li Liu is with the College of System Engineering, National University of Defense Technology, Changsha, 410073, China (e-mail: 
			liuli\_nudt@nudt.edu.cn).\par
			Corresponding author: Yulan Guo (email: yulan.guo@nudt.edu.cn).}}

	\markboth{Journal of \LaTeX\ Class Files,~Vol.~XX, No.~XX, JANUARY~2019}%
	{Shell \MakeLowercase{\textit{et al.}}: Bare Demo of IEEEtran.cls for IEEE Transactions on Magnetics Journals}

	\maketitle

	\IEEEdisplaynontitleabstractindextext{
		\begin{abstract}
			Video super-resolution (SR) aims at generating a sequence of high-resolution (HR) frames with plausible and temporally consistent details from their low-resolution (LR) counterparts. The key challenge for video SR lies in the effective exploitation of temporal dependency between consecutive frames. Existing deep learning based methods commonly estimate optical flows between LR frames to provide temporal dependency. However, the resolution conflict between LR optical flows and HR outputs hinders the recovery of fine details. 
			In this paper, we propose an end-to-end video SR network to super-resolve both optical flows and images. Optical flow SR from LR frames provides accurate temporal dependency and ultimately improves video SR performance. Specifically, we first propose an optical flow reconstruction network (OFRnet) to infer HR optical flows in a coarse-to-fine manner. Then, motion compensation is performed using HR optical flows to encode temporal dependency. Finally, compensated LR inputs are fed to a super-resolution network (SRnet) to generate SR results. Extensive experiments have been conducted to demonstrate the effectiveness of HR optical flows for SR performance improvement. Comparative results on the Vid4 and DAVIS-10 datasets show that our network achieves the state-of-the-art performance.
		\end{abstract}
		
	}

	%
	\IEEEpeerreviewmaketitle

	\section{Introduction}
	%
	%
	%
	%
	\IEEEPARstart{S}{uper}-resolution (SR) aims at generating high-resolution (HR) images from their low-resolution (LR) counterparts. As a typical low-level computer vision problem, SR has been investigated for decades \cite{2001-AComputationallyEfficientSuperresolutionImageReconstructionAlgorithm-Nguyen-573-583,2007-ImageUpsamplingViaImposedEdgeStatistics-Fattal-95-95,2011-ImageandVideoUpscalingfromLocalSelfExamples-Freedman-1-11}. Recently, converting LR videos into HR ones, namely video SR, is under great demand due to the prevalence of high-definition displays. Compared to a single image, adjacent frames in a video clip provide additional information for SR. Therefore, exploiting temporal dependency between consecutive frames plays an important role in video SR.
	
	To exploit temporal dependency between consecutive frames, traditional video SR (or multi-image SR) methods detect recurrent patches across images using patch similarities \cite{2009-GeneralizingtheNonlocalMeanstoSuperResolutionReconstruction-Protter-36-51,2009-SuperResolutionwithoutExplicitSubpixelMotionEstimation-Takeda-1958-1975}. However, these methods can only employ pixel-level dependency and their computational cost is high. To employ sub-pixel dependency, several methods have been proposed to use sub-pixel motion information through optical flow estimation   \cite{2007-OpticalFlowBasedSuperResolution:aProbabilisticApproach-Fransens-106-115,2014-OnBayesianAdaptiveVideoSuperResolution-Liu-346-360,2015-HandlingMotionBlurinMultiFrameSuperResolution-Ma-5224-5232}. These methods formulate the video SR task as an optimization problem and estimate HR images, optical flows and blur kernels alternately. Since a large number of iterations are required to reach convergence, these methods also suffer from high computational costs. 
	
	Motivated by the success of deep learning in single image SR \cite{2014-LearningaDeepConvolutionalNetworkforImageSuperResolution-Dong-184-199,2016-DeeplyRecursiveConvolutionalNetworkforImageSuperResolution-Kim-1637-1645,2017-DeepLaplacianPyramidNetworksforFastandAccurateSuperResolution-Lai-5835-5843}, numerous deep learning based video SR methods have been proposed recently \cite{2015-VideoSuperResolutionViaDeepDraftEnsembleLearning-Liao-531-539,2016-VideoSuperResolutionwithConvolutionalNeuralNetworks-Kappeler-109-122,2017-RealTimeVideoSuperResolutionwithSpatioTemporalNetworksandMotionCompensation-Caballero-2848-2857}. These methods first estimate optical flows from LR frames for motion compensation, and then learn a direct mapping from compensated LR frames to the HR output. Motion compensation encodes temporal dependency in compensated LR frames and facilitates these methods to exploit temporal information from consecutive frames. However, the accuracy of temporal dependency provided by LR optical flows is still low for video SR \cite{2013-SimultaneousSuperResolutionofDepthandImagesUsingaSingleCamera-Lee-281-288}, especially for scenarios with large upscaling factors. 
	
	Since video SR aims at generating high-quality videos with plausible and temporally consistent details, both temporal details and spatial details are important for video SR.  Although existing deep learning based video SR methods \cite{2015-VideoSuperResolutionViaDeepDraftEnsembleLearning-Liao-531-539,2016-VideoSuperResolutionwithConvolutionalNeuralNetworks-Kappeler-109-122,2017-RealTimeVideoSuperResolutionwithSpatioTemporalNetworksandMotionCompensation-Caballero-2848-2857} can successfully hallucinate spatial details from consecutive LR frames, the restoration of temporal details is still under investigated. To address this limitation, we use a convolutional neural network (CNN) to recover HR temporal details in LR frames for video SR. 
	
	In this paper, we propose an end-to-end network to Super-resolve Optical Flows for Video SR (namely, SOF-VSR). Our SOF-VSR network can recover temporal details through optical flow SR, which improves both the accuracy and consistency of video SR. Specifically, we first propose an optical flow reconstruction net (OFRnet) to reconstruct HR optical flows in a coarse-to-fine manner. Different from previous methods \cite{2015-VideoSuperResolutionViaDeepDraftEnsembleLearning-Liao-531-539,2017-RealTimeVideoSuperResolutionwithSpatioTemporalNetworksandMotionCompensation-Caballero-2848-2857,2017-DetailRevealingDeepVideoSuperResolution-Tao-4482-4490} that use optical flows to align LR frames, our OFRnet learns to infer HR optical flows to align latent HR frames. These HR optical flows are then used to perform motion compensation on LR frames. Meanwhile, a space-to-depth transformation is used to bridge the resolution gap between HR optical flows and LR frames.  Finally, these compensated LR frames are fed to a super-resolution net (SRnet) to generate an HR frame. Ablation study is performed to test the effectiveness of HR optical flows for SR performance improvement. Comparative results show that our SOF-VSR network achieves the state-of-the-art performance on the Vid4 and DAVIS-10 datasets.

	The major contributions of our work can be summarized as follows: 
	
	\begin{itemize}
		\item
		We incorporate the SR of both optical flows and images into a unified SOF-VSR network. The SR of optical flows contributes to the SR of images. Consequently, better performance can be achieved by our SOF-VSR network. 
		\item
		We propose an OFRnet to infer HR optical flows from LR frames in a coarse-to-fine manner. It is demonstrated that OFRnet can recover accurate temporal details for SR performance improvement.
		\item
		Our SOF-VSR network achieves the state-of-the-art performance as compared to recent video SR methods.
	\end{itemize}
	
	This work is an extension of our previous conference version \cite{2018-LearningforVideoSuperResolutionthroughHROpticalFlowEstimation-LongguangWang--} with four notable improvements. \textbf{First}, we introduce a more lightweight and compact architecture for SOF-VSR in this paper. Specifically, techniques including channel split, channel shuffle and depth-wise convolution \cite{Ma2018a} are employed to update our building blocks, and the OFRnet is rebuilt using a scale-recurrent network. Our lightweight SOF-VSR network achieves comparable performance to the original one \cite{2018-LearningforVideoSuperResolutionthroughHROpticalFlowEstimation-LongguangWang--}  with parameters being reduced by over 30\%. \textbf{Second}, we have included additional analyses on the design of our network, including ablation studies on HR optical flows, scale-recurrent architecture and building block. \textbf{Third}, we have conducted additional experiments on different upscaling factors and performed additional evaluation on computational complexity. \textbf{Fourth}, additional experiments have been provided to further test the video SR performance through a face recognition task.

	The rest of this paper is organized as follows. In Section II, we briefly review the related works. In Section III, we describe the proposed network in details. In Section IV, experimental results are presented. Finally, we conclude this paper in Section V. 
	
	\section{Related Work}
	In this section, we briefly review several methods that are closely related to our work.
	
	\subsection{Single Image SR}
	Interpolation-based approaches (\emph{e.g.}, bilinear, bicubic and Lanczos \cite{1990-FiltersforCommonResamplingTasks-Turkowski-147-165}) are initially used to increase the size of a single image. However, these methods cannot recover high-frequency details \cite{2003-SuperResolutionImageReconstruction:aTechnicalOverview-Park-21-36}. Later, numerous reconstruction-based approaches have been proposed for single image SR \cite{2005-ImageupSamplingUsingTotalVariationRegularizationwithaNewObservationModel-Aly-1647-1659,2007-ImageUpsamplingViaImposedEdgeStatistics-Fattal-95-95,2008-ImageSuperResolutionUsingGradientProfilePrior-Sun--}. These methods formulate the single image SR task as an optimization problem and introduce different regularization techniques to reconstruct HR images. However, these methods require a large number of iterations and thus suffer from a very high computational cost. To learn a direct mapping between LR and HR images, exemplars are collected from the input image \cite{2009-SuperResolutionfromaSingleImage-Glasner-349-356,2011-ImageandVideoUpscalingfromLocalSelfExamples-Freedman-1-11} and external datasets \cite{2012-CoupledDictionaryTrainingforImageSuperResolution-Yang-3467-3478,2013-AnchoredNeighborhoodRegressionforFastExampleBasedSuperResolution-Timofte-1920-1927}. These exemplar-based methods usually use machine learning approaches (\emph{e.g.}, Markov random field) to achieve promising performance \cite{2014-SingleImageSuperResolution:aBenchmark-Yang-372-386}. For comprehensive reviews on traditional single image SR methods, we refer the readers to \cite{2003-SuperResolutionImageReconstruction:aTechnicalOverview-Park-21-36,2014-SingleImageSuperResolution:aBenchmark-Yang-372-386}.
	
Recently, deep learning has been extensively investigated for SR. Dong \emph{et al.} \cite{2014-LearningaDeepConvolutionalNetworkforImageSuperResolution-Dong-184-199} proposed the pioneering work to use deep learning for single image SR. They used a three-layer CNN (namely, SRCNN) to approximate the non-linear mapping from an LR image to its corresponding HR image. Kim \emph{et al.} \cite{2016-AccurateImageSuperResolutionUsingVeryDeepConvolutionalNetworks-Kim-1646-1654} proposed a very deep super-resolution network (i.e., VDSR) with 20 convolutional layers. The deep architecture of VDSR improves the approximating capacity of CNN to achieve better performance. To achieve a compromise between model size and SR performance, Tai \emph{et al.} \cite{2017-ImageSuperResolutionViaDeepRecursiveResidualNetwork-Tai-2790-2798} developed a deep recursive residual network (DRRN) to deepen the network without obvious increase in model parameters. Shi \emph{et al.} \cite{2016-RealTimeSingleImageandVideoSuperResolutionUsinganEfficientSubPixelConvolutionalNeuralNetwork-Shi-1874-1883} proposed an efficient sub-pixel convolutional neural network (ESPCN) to increase the resolution of an LR image at the end of the network. Its computational complexity is significantly reduced. More recently, Zhang \emph{et al.} \cite{2018-ResidualDenseNetworkforImageSuperResolution-Zhang--} proposed a residual dense network (RDN) to facilitate effective feature learning using a contiguous memory mechanism.
	
	\subsection{Video SR}
	\subsubsection{Traditional Video SR}
	Since the seminal work proposed by Tsai and Huang \cite{1984-MultiframeImageRestorationandRegistration-Tsai--}, significant progresses have been achieved in multi-image SR and  video SR. Early methods
	\cite{1996-ExtractionofHighResolutionFramesfromVideoSequences-Schultz-996-1011,1997-JointMAPRegistrationandHighResolutionImageEstimationUsingaSequenceofUndersampledImages-Hardie-1621-1633} focus on videos with only affine transforms exist between adjacent frames, which is usually not the real case. To handle complex motion patterns in video clips, Protter \emph{et al.} \cite{2009-GeneralizingtheNonlocalMeanstoSuperResolutionReconstruction-Protter-36-51} generalized the non-local means framework for video SR. They performed adaptive fusion of multiple frames using patch-wise spatio-temporal similarities. Takeda \emph{et al.} \cite{2009-SuperResolutionwithoutExplicitSubpixelMotionEstimation-Takeda-1958-1975} further
	introduced a 3D kernel regression to exploit patch-wise spatio-temporal neighborhood
	relationship. However, HR images produced by these two methods are usually over-smoothed. To exploit pixel-wise correspondences, optical flow estimation was used in \cite{2007-OpticalFlowBasedSuperResolution:aProbabilisticApproach-Fransens-106-115,2014-OnBayesianAdaptiveVideoSuperResolution-Liu-346-360,2015-HandlingMotionBlurinMultiFrameSuperResolution-Ma-5224-5232}. These methods formulate the video SR task as an optimization problem and use iterative frameworks to estimate HR images, optical flows and blur kernels alternately. However, these methods are time-consuming.
	
	\subsubsection{Deep Video SR with Separated Motion Compensation}
	Inspired by the success of SRCNN in single image SR, deep learning has been investigated for video SR. Kappelar \emph{et al.} \cite{2016-VideoSuperResolutionwithConvolutionalNeuralNetworks-Kappeler-109-122} proposed a two-step framework to perform video SR. Specifically, optical flow estimation is first performed for motion compensation. Then, the compensated frames are concatenated and fed to a CNN to reconstruct an HR frame. Following the same two-step framework as \cite{2016-VideoSuperResolutionwithConvolutionalNeuralNetworks-Kappeler-109-122}, Liao \emph{et al.} \cite{2015-VideoSuperResolutionViaDeepDraftEnsembleLearning-Liao-531-539} estimated multiple optical flows using different parameter settings. These optical flows are then used for motion compensation to generate an ensemble of SR-drafts. Finally, a CNN is employed to recover high-frequency details from the ensemble. 
	The two-step framework separates motion estimation and compensation from the CNN network. Therefore, it is difficult for these methods to obtain an overall optimal solution.
	
	\begin{figure*}[ht]
		\centering
		\includegraphics[width=0.9\linewidth]{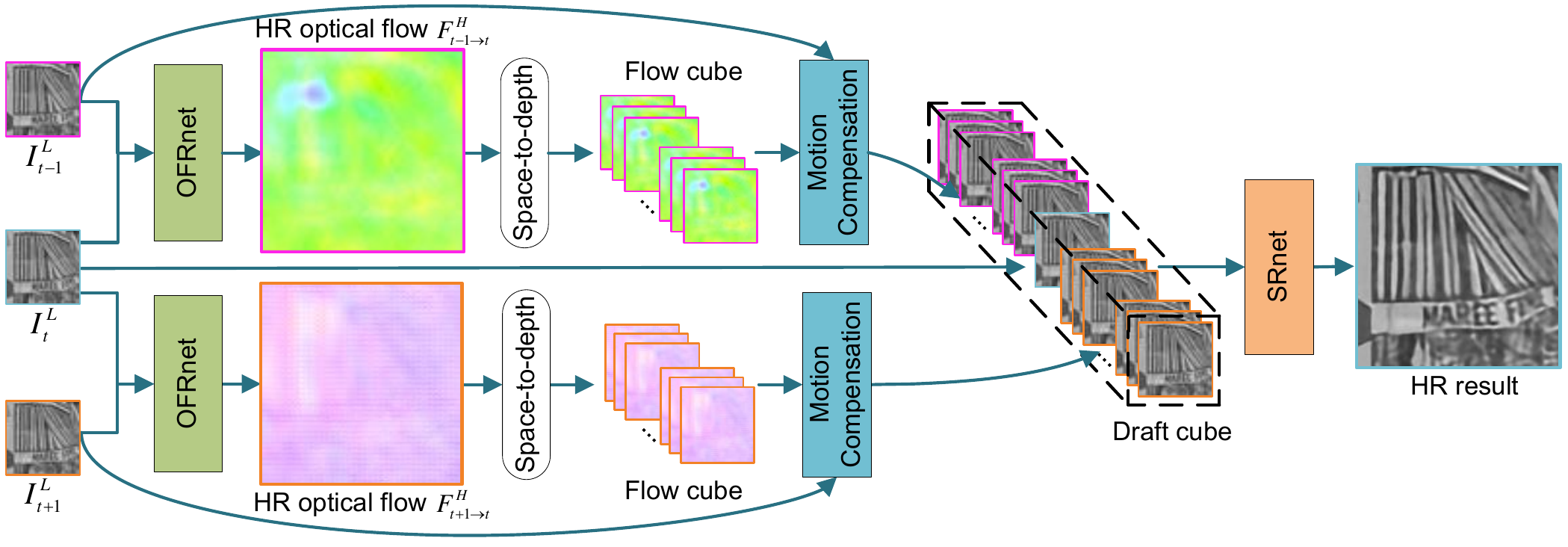}
		\caption{An overview of our SOF-VSR network. Our network is fully convolutional and can be trained in an end-to-end manner.}
		\label{fig1}
	\end{figure*}
	
	\subsubsection{Deep Video SR with Integrated Motion Compensation} 
	Recently, Caballero \emph{et al.} \cite{2017-RealTimeVideoSuperResolutionwithSpatioTemporalNetworksandMotionCompensation-Caballero-2848-2857} proposed the first end-to-end CNN (namely, VESPCN) for video SR to integrate both motion estimation and compensation. Their VESPCN network comprises a motion estimation module and a spatio-temporal ESPCN module \cite{2016-RealTimeSingleImageandVideoSuperResolutionUsinganEfficientSubPixelConvolutionalNeuralNetwork-Shi-1874-1883}. Since then, end-to-end framework with integrated motion compensation
	dominates the research of video SR. 
	Tao \emph{et al.} \cite{2017-DetailRevealingDeepVideoSuperResolution-Tao-4482-4490}
	used the motion estimation module in VESPCN and then designed a new layer to achieve both sub-pixel motion compensation (SPMC) and resolution enhancement. They also proposed an encode-decoder network with LSTM to learn temporal contexts.
	Liu \emph{et al.} \cite{2017-RobustVideoSuperResolutionwithLearnedTemporalDynamics-Liu--} customized ESPCN \cite{2016-RealTimeSingleImageandVideoSuperResolutionUsinganEfficientSubPixelConvolutionalNeuralNetwork-Shi-1874-1883} to
	simultaneously reconstruct HR frames using different numbers of LR frames. A temporal adaptive network (namely, TDVSR) is then introduced to aggregate multiple HR estimates with learned dynamic weights. 
	Sajjadi \emph{et al.} \cite{2018-FrameRecurrentVideoSuperResolution-Sajjadi-6626-6634} proposed a frame-recurrent architecture (namely, FRVSR) to use previously inferred HR estimates for the SR of subsequent frames. This recurrent architecture can assimilate previous inferred HR frames without increasing computational costs.
	
	\subsubsection{Deep Video SR without Explicit Motion Compensation} 
	Huang \emph{et al.} \cite{2017-VideoSuperResolutionandViaBidirectionalandRecurrentConvolutionalandNetworks-Huang-1-1} proposed a bidirectional recurrent CNN to avoid explicit motion estimation and compensation. This recurrent-like architecture can capture long-term contextual information within temporal sequences. However, this method fails to handle large displacements and other complicated motions. Jo \emph{et al.} \cite{2018-DeepVideoSuperResolutionNetworkUsingDynamicUpsamplingFilterswithoutExplicitMotionCompensation-Jo--} introduced a CNN to generate dynamic upsampling filters for video SR. These dynamic upsampling filters are computed using local spatio-temporal neighborhood to avoid explicit motion compensation.
	
	Since temporal dependency between consecutive frames is important for video SR, existing deep learning based video SR methods focus on explicit or implicit exploitation of temporal dependency. However, these methods model temporal dependency in LR space, their limited accuracy in dependency hinders the restoration of fine details. Different from previous works, we propose an end-to-end video SR network to recover both temporal details and spatial details. Specifically, we first super-resolve optical flows to recover temporal details. These HR optical flows provide accurate temporal dependency and contribute to the restoration of spatial details. It is demonstrated that optical flow SR facilitates our network to achieve the state-of-the-art performance.
	
	\section{Methodology}
	In this section, we introduce our SOF-VSR network in details. We first give an overview of our SOF-VSR network, and then describe the OFRnet, the motion compensation module and the SRnet of our network. Finally, we present the loss function for the training of our network.
	
	\subsection{Overview}
	Given $T$ consecutive LR frames ($I_{t-N}^{L},...,I_{t}^{L},...,I_{t+N}^{L}$) of a video clip as the input of SOF-VSR, our task is to super-resolve the central frame. Here, $T\!=\!2N+1$. Following \cite{2017-RobustVideoSuperResolutionwithLearnedTemporalDynamics-Liu--}, we convert input LR frames into YCbCr color space and only process the luminance channel. Input LR frames are first fed to OFRnet to infer HR optical flows. Specifically, our OFRnet takes  the central LR frame $I_{t}^{L}$ and one neighboring frame $I_{i}^{L}$ as input to generate an HR optical flow $F_{i\rightarrow{t}}^{H}$. Then, a space-to-depth
	transformation \cite{2018-FrameRecurrentVideoSuperResolution-Sajjadi-6626-6634}
	is employed to shuffle the HR optical flows into LR grids, resulting in LR flow cubes. Next, motion compensation is performed to generate a draft cube using these flow cubes. Finally, the draft cube is fed to SRnet to infer the HR frame. The overview of
	our network is shown in Fig. \ref{fig1}. For simplicity, we only show the architecture with $T=3$.
	
	\subsection{Optical Flow Reconstruction Net (OFRnet)}
	
	\begin{figure*}[ht]
		\centering
		\includegraphics[width=0.99\linewidth]{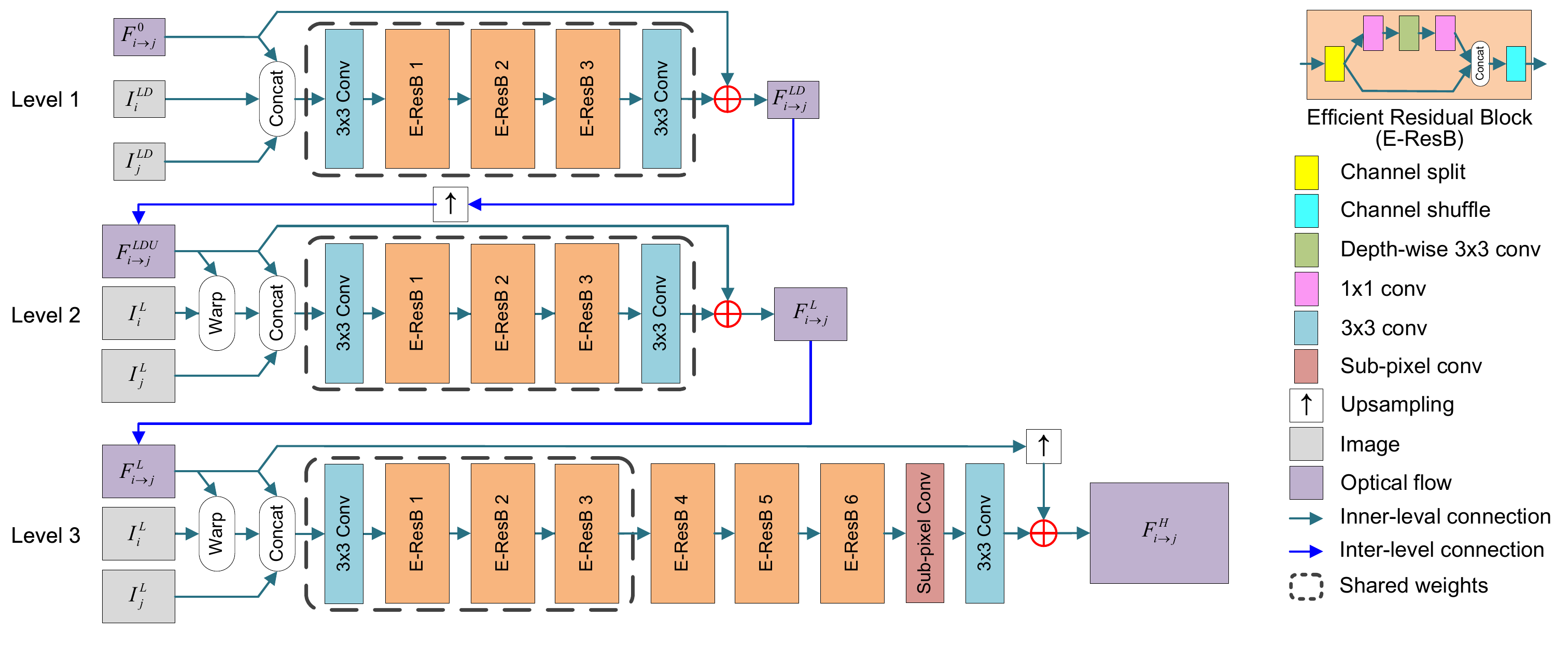}
		\caption{\textcolor{black}{The architecture of our OFRnet. Our OFRnet works in a coarse-to-fine manner. At each level, the output of its previous level is used to generate a residual optical flow.}}
		\label{fig2}
	\end{figure*}
	
	It has already been demonstrated by deep learning based SR methods (\emph{e.g.}, SRCNN \cite{2014-LearningaDeepConvolutionalNetworkforImageSuperResolution-Dong-184-199}, VDSR \cite{2016-AccurateImageSuperResolutionUsingVeryDeepConvolutionalNetworks-Kim-1646-1654} and RDN \cite{2018-ResidualDenseNetworkforImageSuperResolution-Zhang--})  that CNN is able to learn the non-linear mapping between LR and HR images. Recent CNN-based optical flow estimation methods (\emph{e.g.}, FlowNet \cite{2015-FlowNet:LearningOpticalFlowwithConvolutionalNetworks-Dosovitskiy-2758-2766}, PWCNet \cite{2017-PWCNet:CNNsforOpticalFlowUsingPyramidWarpingandCostVolume-Sun--} and LiteFlowNet \cite{2018-LiteFlowNet:aLightweightConvolutionalNeuralNetworkforOpticalFlowEstimation-Hui--}) have also shown the potential for motion estimation. Therefore, we incorporate these two tasks into a unified OFRnet to infer HR optical flows directly from LR images. Specifically, our OFRnet takes a pair of LR frames $I_{i}^{L}$ and $I_{j}^{L}$
	as inputs, and reconstruct an optical flow $F_{i\rightarrow j}^{H}$ between their corresponding
	HR frames $I_{i}^{H}$ and $I_{j}^{H}$:
	
	\begin{equation}
	F_{i\rightarrow j}^{H}=\mathbf{Net}_{OFR}(I_{i}^{L},\,I_{j}^{L};\,\Theta_{OFR}),
	\end{equation}
	where $F_{i\rightarrow j}^{H}$ represents the HR optical flow and $\Theta_{OFR}$ denotes
	the set of parameters.
	
	Multi-scale mechanism has been demonstrated to be effective in optical flow estimation \cite{2017-OpticalFlowEstimationUsingaSpatialPyramidNetwork-Ranjan-2720-2729,2018-LiteFlowNet:aLightweightConvolutionalNeuralNetworkforOpticalFlowEstimation-Hui--}, stereo matching \cite{2017-EndtoEndLearningofGeometryandContextforDeepStereoRegression-Kendall-66-75,2018-PyramidStereoMatchingNetwork-Chang--} and many other vision tasks \cite{2018-ScaleRecurrentNetworkforDeepImageDeblurring-Tao--}. To reduce model size and training difficulty, a scale-recurrent architecture with shared parameters across scales is used in SRN-DeblurNet \cite{2018-ScaleRecurrentNetworkforDeepImageDeblurring-Tao--}. Inspired by this, we introduce a scale-recurrent network for optical flow reconstruction, as illustrated in Fig. \ref{fig2}. For the first two levels, we use a recurrent module to estimate optical flows for inputs with different scales. For level 3, we first use the recurrent structure to generate deep representations, and then introduce an SR module to recover HR optical flows from the LR feature representations. The scale-recurrent architecture enables  OFRnet to handle complex motion patterns (especially large displacements) while being lightweight and compact.
	
	\textbf{Level 1:} The pair of input LR images $I_{i}^{L}$ and $I_{j}^{L}$ are first downsampled by a factor of 2 to produce $I_{i}^{LD}$ and $I_{j}^{LD}$. Meanwhile, an initial flow map $F_{i\rightarrow j}^{0}$ with all elements of 0 is generated. The initial flow map $F_{i\rightarrow j}^{0}$ is concatenated with $I_{i}^{LD}$ and $I_{j}^{LD}$ and then fed to a feature extraction layer with 320 kernels of size $3\times3$. Then, three efficient residual blocks are used to generate deep features. Channel split, channel shuffle and depth-wise convolution techniques  \cite{Ma2018a} are used in these residual blocks to improve the efficiency. Next, these features are fed to a flow estimation layer with 2 kernels of size $3\times3$ to generate optical flow  $F_{i\rightarrow j}^{LD}$ at this level. All convolutional layers are followed by a leaky rectified linear unit (ReLU) except the middle layer in each residual block and the last flow estimation layer.
	
	\textbf{Level 2:} Once the optical flow $F_{i\rightarrow j}^{LD}$
	is obtained from level 1, it is upscaled by a factor
	of 2 using bilinear interpolation. Note that, the magnitude of optical flow is also doubled with the resolution. The upscaled flow $F_{i\rightarrow j}^{LDU}$ is then used to warp $I_{i}^{L}$, resulting in ${I}_{i\rightarrow j}^{L}$. Next, ${I}_{i\rightarrow j}^{L}$, $I_{j}^{L}$ and $F_{i\rightarrow j}^{LDU}$ are concatenated and fed
	to the recurrent module (which is the same as the one used in level 1) to generate optical flow  $F_{i\rightarrow j}^{L}$ at this level.
	
	\textbf{Level 3:} Since the output optical flow $F_{i\rightarrow j}^{L}$ of level 2 has the same size as the LR input $I_{j}^{L}$, level 3 works as an SR module to reconstruct HR optical flows. Similar to level 2, ${I}_{i\rightarrow j}^{L}$, $I_{j}^{L}$ and $F_{i\rightarrow j}^{L}$ are first concatenated and fed to the recurrent module (which is the same as the one used in levels 1 and 2) to extract features. These features are then fed to three additional  residual blocks to generate deep representations. Next, the resulting feature representations are fed to a sub-pixel layer \cite{2016-RealTimeSingleImageandVideoSuperResolutionUsinganEfficientSubPixelConvolutionalNeuralNetwork-Shi-1874-1883}  for resolution enhancement. Finally, a flow estimation layer is used to generate the final HR optical flow $F_{i\rightarrow j}^{H}$.
	
	Although numerous networks for SR \cite{2016-RealTimeSingleImageandVideoSuperResolutionUsinganEfficientSubPixelConvolutionalNeuralNetwork-Shi-1874-1883,2017-DeepLaplacianPyramidNetworksforFastandAccurateSuperResolution-Lai-5835-5843,2018-DeepBackProjectionNetworksforSuperResolution-Haris--} and optical flow estimation \cite{2015-FlowNet:LearningOpticalFlowwithConvolutionalNetworks-Dosovitskiy-2758-2766,2017-PWCNet:CNNsforOpticalFlowUsingPyramidWarpingandCostVolume-Sun--,2018-LiteFlowNet:aLightweightConvolutionalNeuralNetworkforOpticalFlowEstimation-Hui--} can be found in literature, our OFRnet is, to the best of our knowledge, the first unified network to integrate these two tasks. Specifically, our OFRnet learns to infer HR optical flows between latent HR images from LR inputs. Though some existing video SR methods can also obtain optical flows of full resolution by performing interpolation on LR inputs \cite{2017-EndtoEndLearningofVideoSuperResolutionwithMotionCompensation-Makansi-203-214} or LR optical flows \cite{2018-FrameRecurrentVideoSuperResolution-Sajjadi-6626-6634}, their flow estimation is still performed in LR space since interpolation does not introduce additional information for SR \cite{2016-RealTimeSingleImageandVideoSuperResolutionUsinganEfficientSubPixelConvolutionalNeuralNetwork-Shi-1874-1883}. Note that, inferring HR optical flows from LR images is quite challenging, our OFRnet has demonstrated the potential of CNN to address this challenge. It is further demonstrated in Sec. \ref{sec3.3} that our SOF-VSR network is benefited from HR optical flows in terms of both accuracy and consistency.
	
	\subsection{Motion Compensation Module}
	
	\begin{figure}[bt]
		\centering
		\includegraphics[width=0.9\linewidth]{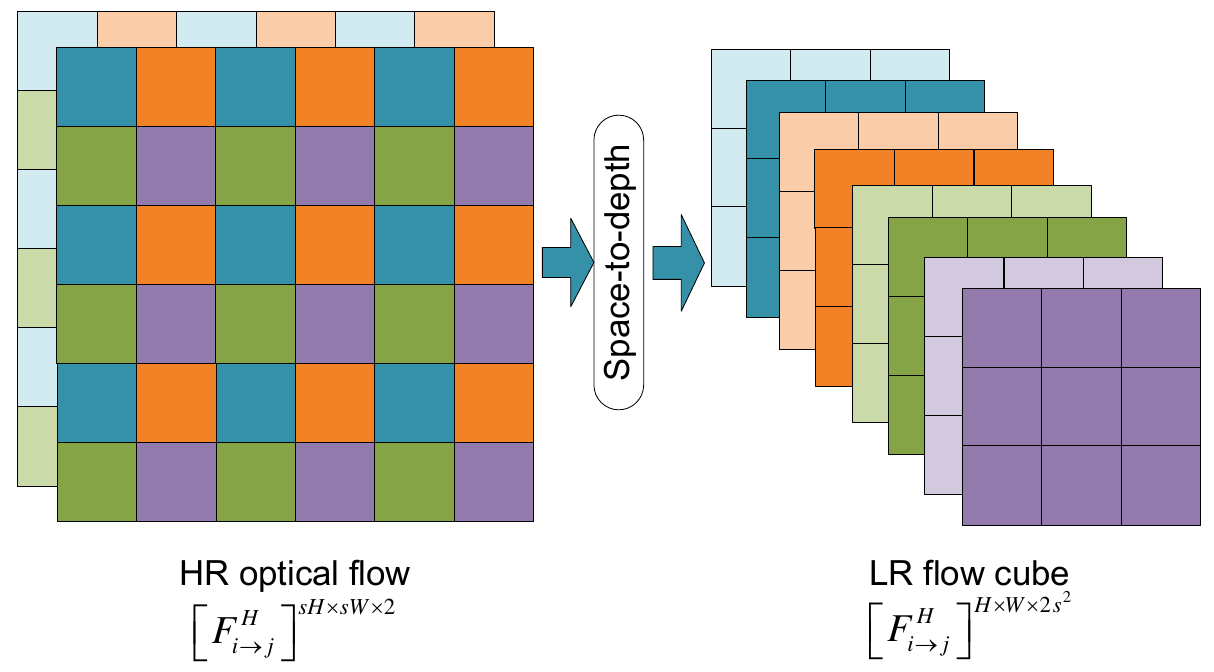}
		\caption{An illustration of space-to-depth transformation. The space-to-depth transformation folds an HR optical flow in LR space to generate an LR flow cube.}
		\label{fig3}
	\end{figure}
	
	Once HR optical flows are produced by OFRnet, space-to-depth transformation is used to bridge the resolution gap between HR optical flows and
	LR frames. As shown in Fig. \ref{fig3}, regular LR grids are extracted
	from the HR flow and placed into the channel dimension to derive
	a flow cube with the same resolution as LR frames:
	\begin{equation}
	\left[F_{i\rightarrow j}^{H}\right]^{sH\times sW\times2}\rightarrow\left[F_{i\rightarrow j}^{H}\right]^{H\times W\times2s^{2}},
	\end{equation}
	where $H$ and $W$ represent the size of the LR frame, $s$ is the upscaling factor. Note that, the magnitude of optical flow is divided by a scalar $s$ during the transformation to match the spatial resolution of LR frames.
	
	Then, slices are extracted from the LR flow
	cube to warp the LR frame $I_{i}^{LR}$, resulting in multiple warped drafts:
	\begin{equation}
	C_{i\rightarrow j}^{L}=\textup{W}(I_{i}^{L},\,\left[F_{i\rightarrow j}^{H}\right]^{H\times W\times2s^{2}}),
	\end{equation}
	where $\textup{W}(\cdot)$ denotes the warping operation using bilinear interpolation and $C_{i\rightarrow j}^{L}\!\in\!R^{H\times W\times{s^{2}}}$ represents
	the concatenation of multiple warped drafts. Note that, although motion compensation is performed on LR frames, accurate temporal dependency can be encoded in compensated frames since HR optical flows are employed.
	
	\subsection{Super-Resolution Net (SRnet)}
	
	Our SOF-VSR takes $T$ consecutive LR frames ($I_{t-N}^{L},...,I_{t}^{L},...,I_{t+N}^{L}$) as inputs to super-resolve the central frame. After motion compensation, multiple drafts are produced for each neighboring frame. As shown in Fig. \ref{fig1}, all the drafts are concatenated with the central LR frame and fed to SRnet to infer the HR frame:
	\begin{equation}
	I_0^{SR}=\mathbf{Net}_{SR}(C^{L};\,\Theta_{SR}),
	\end{equation}
	where $I_{0}^{SR}$ is the SR result of the central frame and $\Theta_{SR}$ is the set of parameters. $C^{L}\!\in\!R^{H\times W\times(2Ns^{2}+1)}$ represents the concatenation of all drafts after motion compensation, namely, draft cube.

\begin{figure}[bt]
		\centering
		\includegraphics[width=0.95\linewidth]{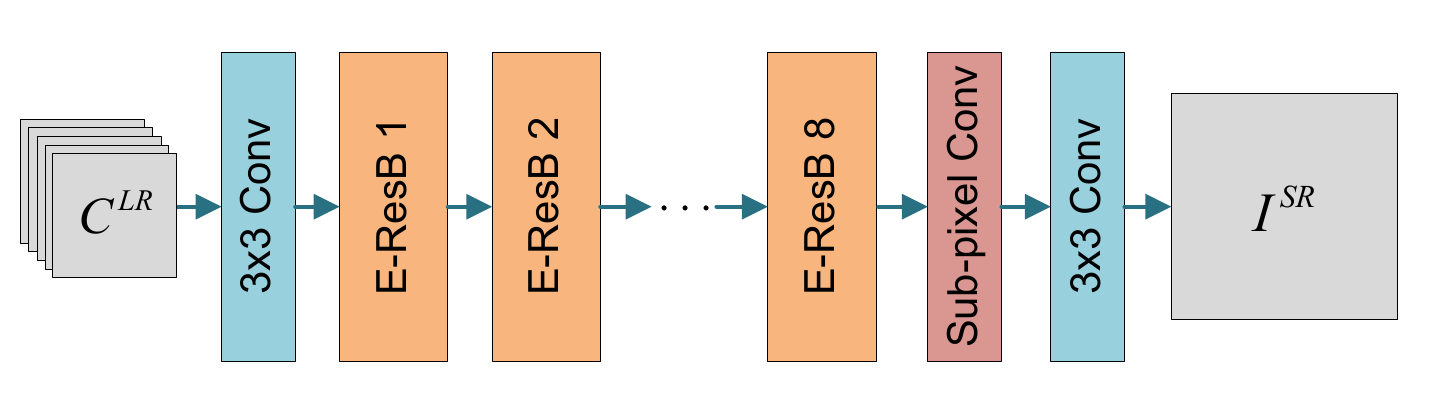}
		\caption{The architecture of our SRnet.}
		\label{fig4}
	\end{figure}

	\textcolor{black}{As shown in Fig. \ref{fig4}, the draft cube is first passed to a feature extraction layer with 320 kernels of size $3\times3$ for feature extraction. The output features are then fed to 8 efficient residual  blocks to generate deep features. Once features are generated by these residual blocks, they are fed to a sub-pixel layer for resolution enhancement. Finally, a $3\times3$ convolutional layer is used to generate the HR frame. Since our SOF-VSR network only works on the luminance channel, the number of kernels in the last layer is set to 1.}
	
	\subsection{Loss Function}
	
	We design two loss terms $\mathcal{L_{\mathrm{SR}}}$ and $\mathcal{L_{\mathrm{OFR}}}$ for SRnet and OFRnet, respectively. For the training of SRnet, we use the mean square error (MSE) loss:
	
	\begin{equation}
	\mathcal{L_{\mathrm{SR}}}=\left\Vert I_0^{SR}-I_0^{H}\right\Vert _{2}^{2}.
	\end{equation}
	
	For the
	training of OFRnet, intermediate supervision is used at each level
	of the pyramid:
	\begin{equation}
	\mathcal{L_{\mathrm{OFR}}}\!=\!\sum_{i\in[-N,\,N],\,i\neq0}\!\frac{
		\mathcal{L}_{{level 3},i}\!+\!\lambda_{2}\mathcal{L}_{{level 2},i}+\!\lambda_{1}
		\mathcal{L}_{{level 1},i}}{2N},
	\end{equation}
	where
	\begin{equation}
	\left\{
	\begin{aligned}
	\mathcal{L}_{{level 3},i}&\!=\! \left\Vert \textup{W}(I_{i}^{H},\,F_{i\rightarrow0}^{H})\!-\!I_{0}^{H}
	\right\Vert _{1}\!+\!\lambda_{3} \left\Vert \nabla F_{i\rightarrow0}^{H}\right\Vert _{1} \\
	\mathcal{L}_{{level 2},i}&\!=\! \left\Vert \textup{W}(I_{i}^{L},\,F_{i\rightarrow0}^{L})\!-\!I_{0}^{L}
	\right\Vert _{1}\!+\!\lambda_{3} \left\Vert \nabla F_{i\rightarrow0}^{L}\right\Vert _{1} \\
	\mathcal{L}_{{level 1},i}&\!=\!\left\Vert \textup{W}(I_{i}^{LD}\!,\,F_{i\rightarrow0}^{LD})\!-\!I_{0}^{LD}
	\right\Vert _{1}\!+\!\lambda_{3} \left\Vert \nabla F_{i\rightarrow0}^{LD}\right\Vert _{1}
	\end{aligned},
	\right.
	\label{equ5}
	\end{equation}
	$\left\Vert \nabla F_{i\rightarrow0}^{H}\right\Vert _{1}$, $\left\Vert \nabla F_{i\rightarrow0}^{L}\right\Vert _{1}$ and $\left\Vert \nabla F_{i\rightarrow0}^{LD}\right\Vert _{1}$ are L1 regularization terms to constrain the smoothness of the optical flows at different scales.
	We empirically set $\lambda_{2}=0.2$ and $\lambda_{1}=0.1$ to make our OFRnet focus on the last level. We also set $\lambda_{3}=0.1$ as the regularization coefficient.

	Finally, the total loss for joint training is defined as $\mathcal{\mathcal{L}=L_{\mathrm{SR}}}+\lambda_{4}\mathcal{L_{\mathrm{OFR}}}$,
	where $\lambda_{4}$ is empirically set to 0.01 to balance these two loss terms.
	
	\section{Experiments}
	In this section, we first introduce the datasets and implementation details. Next, ablation study is performed on the Vid4 dataset to test our network. Our SOF-VSR is then compared to the state-of-the-art methods on the Vid4 and DAVIS-10 datasets. Finally, face recognition task is used to further demonstrate the effectiveness of our network for high-level vision tasks.

	\begin{table*}[htp]
		\caption{Comparative results achieved by our network and its variants on the Vid4 dataset for $4\times$ SR. FLOPs is computed based on HR frames with a  resolution of 720p (1280$\times$720).}
		\label{tab1}
		\begin{center}
			\small
			\setlength{\tabcolsep}{0.5mm}{
				\begin{tabular}{|l|c|c|c|c|c|c|c|c|c|c|c|}
					\hline 
					& \multirow{2}{*}{PSNR($\uparrow$)}  & \multirow{2}{*}{SSIM($\uparrow$)} & \multirow{2}{*}{\tabincell{c}{T-MOVIE($\downarrow$)\\($\times10^{-3}$)}} 
					& \multirow{2}{*}{\tabincell{c}{MOVIE($\downarrow$) \\($\times10^{-3}$)}}
					& \multicolumn{3}{c|}{\textcolor{black}{Params.}} & \multicolumn{3}{c|}{\textcolor{black}{FLOPs}} 
					\tabularnewline
					\cline{6-11}
					&&&& & OFRnet & SRnet & Overall & OFRnet & SRnet & Overall
					\tabularnewline
					\hline
					SOF-VSR w/o OFRnet  & 25.70 & 0.753 & 20.03 & 4.47 & - & 0.59M & 0.59M & - & 36.10G & 36.10G
					\tabularnewline
					SOF-VSR w/o sub-pixel conv & 25.85 & 0.765 & 19.69 & 4.41 & 0.38M & 0.59M & 0.97M & 35.19G &36.10G &106.48G
					\tabularnewline
					SOF-VSR w/o sub-pixel conv + upsampling & 25.83 & 0.766 & 19.65 & 4.39 & 0.38M & 0.59M & 0.97M & 35.19G &36.10G &106.48G
					\tabularnewline
					\hline
					\textcolor{black}{SOF-VSR SISR} & 25.96 & 0.772 & 19.32 & 4.24 & - & - & 1.05M & - & - & 1.12T
					\tabularnewline
					\hline
					\textcolor{black}{SOF-VSR w scale-cascaded architecture} & 26.02 & 0.773 & 19.16 &4.23 & 0.74M & 0.59M & 1.33M &36.40G &36.10G &108.90G
					\tabularnewline
					\textcolor{black}{SOF-VSR w vanilla residual blocks} & 26.04 & 0.773 & 19.02 &4.20 & 0.67M & 0.89M & 1.56M & 45.51G & 52.12G & 143.14G 
					\tabularnewline
					\hline
					SOF-VSR &26.00  &0.772 & 19.35 & 4.25 &0.41M &0.59M &1.00M &36.40G &36.10G &108.90G
					\tabularnewline
					\hline
			\end{tabular}}
		\end{center}
	\end{table*}

	\subsection{Datasets}
	For training, we collected 145 1080P HD video clips from the CDVL Database\footnote{www.cdvl.org}. These video clips cover diverse natural and urban scenes. Similar to \cite{2018-DeepVideoSuperResolutionNetworkUsingDynamicUpsamplingFilterswithoutExplicitMotionCompensation-Jo--}, we used 4 video clips including \emph{Coastguard}, \emph{Foreman}, \emph{Garden}, and \emph{Husky} from the Derf's collection\footnote{media.xiph.org/video/derf/} for validation. For fair comparison to the state-of-the-arts, we used the widely
	used Vid4 benchmark dataset to test our method. We also used a subset of the DAVIS dataset \cite{2017-The2017DAVISChallengeonVideoObjectSegmentation-Pont-Tuset--} with 10 video clips for further comparison, which will be referred to as DAVIS-10 in this paper. Note that, each video clip in the test dataset contains 31 consecutive frames, the same as in \cite{2015-VideoSuperResolutionViaDeepDraftEnsembleLearning-Liao-531-539}.
	
	\subsection{Implementation Details}
	
	Following  \cite{2015-VideoSuperResolutionViaDeepDraftEnsembleLearning-Liao-531-539,2017-DetailRevealingDeepVideoSuperResolution-Tao-4482-4490}, we downsampled the original video clips to the size of $540\times960$ as the HR groundtruth using Matlab function $imresize$  in bicubic mode. These HR videos were further downsampled to generate LR video clips with different upscaling factors. During the training phase,
	we randomly extracted $T$ consecutive frames from an LR video clip, and randomly cropped a $32\times32$ patch as the input. Meanwhile, its corresponding patch in the HR video clip was cropped as the groundtruth. Data augmentation was performed through rotation and reflection to improve the generalization capability of our network. 
	
	For evaluation, we used peak signal-to-noise ratio (PSNR) and structural similarity index (SSIM) to test the accuracy of each individual frame. \textcolor{black}{The overall PSNR/SSIM values were then calculated by aggregating PSNRs/SSIMs over all frames in a video clip.} To test the consistency performance, we used the temporal motion-based video integrity evaluation index (T-MOVIE) \cite{2010-MotionTunedSpatioTemporalQualityAssessmentofNaturalVideos-Seshadrinathan-335-350}. Moreover, MOVIE \cite{2010-MotionTunedSpatioTemporalQualityAssessmentofNaturalVideos-Seshadrinathan-335-350} was used to test the overall quality of a video. This metric is correlated to human perception and has been widely applied in video quality assessment. All metrics are computed in the luminance channel. Following \cite{2017-NTIRE2017ChallengeonSingleImageSuperResolution:MethodsandResults-Timofte-1110-1121}, borders of $6+s$ are cropped for fair comparison. 
	
	Our SOF-VSR was implemented in PyTorch on a PC with an Nvidia GTX 1080Ti GPU. We used the Adam solver \cite{2015-Adam:aMethodforStochasticOptimization-Kingma--}
	with $\beta_{1}=0.9$, $\beta_{2}=0.999$ and a batch size of 32 for training. The initial learning rate was set to $1\!\times\!10^{-3}$ and divided by 10 after every 80K iterations. The training was stopped after 200K iterations since more iterations do not provide further consistent improvement.
	
		\subsection{Analysis of the Network Architecture}
	\label{sec3.3}
	In this section, we present ablation experiments on the Vid4 dataset to analyze the architecture of our SOF-VSR network. All variants in the experiment were retrained following the configuration of the original SOF-VSR network.
	
	\subsubsection{Motion Compensation} 
	To handle complex motion patterns in video sequences, optical flows are used for motion compensation in our network. To test the effectiveness of motion compensation for video SR, we removed the whole OFRnet module and fed LR frames directly to our SRnet. Note that, replicated LR frames were used to match the dimension of the draft cube $C^{L}$. Results achieved on the Vid4 dataset are listed in Table \ref{tab1}.
	
	It can be observed that the performance of our SOF-VSR significantly benefits from motion compensation. If OFRnet is removed, the PSNR/SSIM values are decreased from 26.00/0.772 to 25.70/0.753. Besides, the consistency performance is also degraded, with T-MOVIE value being increased from 19.35 to 20.03. That is because, it is difficult for SRnet to learn the non-linear mapping between LR and HR images under complex motion patterns.

	\subsubsection{LR Flow vs. HR Flow} 
	
	Optical flow SR provides accurate temporal dependency for video SR. To test the effectiveness of HR optical flows, we replaced the sub-pixel convolution at level 3 in our OFRnet with a normal convolution. Then, the resulting LR optical flows were directly used for motion compensation and subsequent processing. To match the dimension of the draft cube, compensated LR frames were also replicated before feeding to SRnet.
	
	It can be observed from Table \ref{tab1} that if LR optical flows were generated for motion compensation, the PSNR/SSIM values are increased to 25.85/0.765. However, the performance is still inferior to our SOF-VSR using HR optical flows. That is because, HR optical flows provide more accurate temporal dependency for performance improvement.
	
	\textcolor{black}{
	\subsubsection{Upsampled Flow vs. Super-resolved Flow} 	
	Optical flow sup-resolution can also be simply achieved by interpolation. However, our OFRnet can recover more accurate optical flow details. To demonstrate this, we replaced the sub-pixel convolution at level 3 in our OFRnet with a normal convolution, and upsampled the resulting LR optical flows using bilinear interpolation. Then, we used the modules in our original network for subsequent processing. From the comparative results shown in Table \ref{tab1}, we can see that if bilinear interpolation is used to upsample LR optical flows, no significant improvement can be observed (25.85/0.765 \emph{vs.} 25.83/0.766). That is because, the upsampling operator cannot recover temporal dependency reliably. If optical flow SR is performed, the PSNR/SSIM values are increased to 26.00/0.772. That is because, optical flow SR can recover finer temporal details and facilitate our SOF-VSR network to achieve better video SR performance.}
	
	\begin{figure*}[t]
		\centering
		\includegraphics[width=1\linewidth]{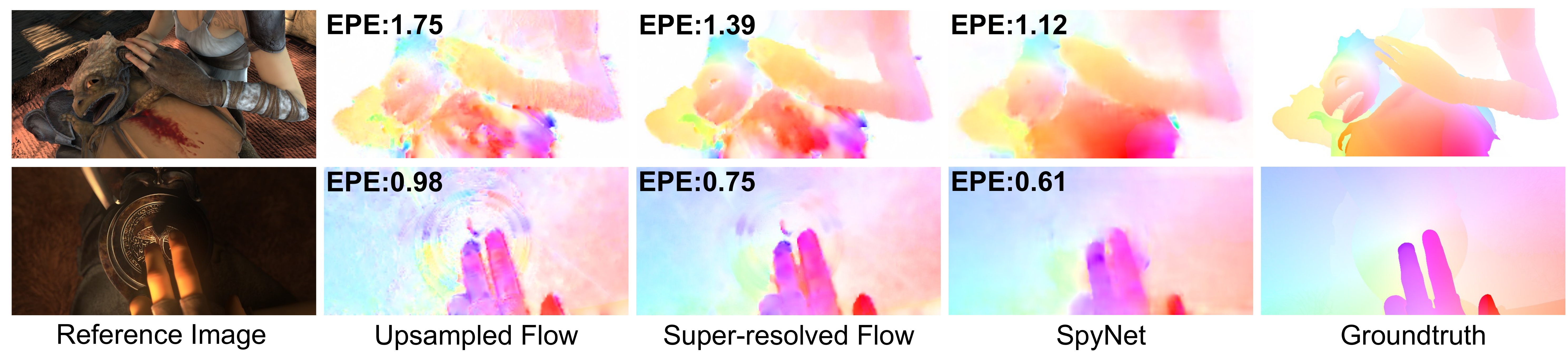}
		\caption{\textcolor{black}{Visual comparison of optical flow estimation results achieved on the Sintel dataset for $4\times$ SR. The super-resolved optical flows recover finer correspondences with more clear edges and fewer artifacts \textcolor{black}{than the upsampled optical flows}.}}
		\label{fig5}
	\end{figure*}
	
	\begin{table}[t]
		\caption{\textcolor{black}{Average EPE results achieved on the training sets of Sintel, Middlebury, KITTI 2012 and KITTI 2015 for $4\times$ SR. Best results are shown in boldface.}}
		\label{tab2}
		\begin{center}
			\small
			\setlength{\tabcolsep}{0.8mm}{
				\begin{tabular}{|l|c|c|c||c|c|}
					\hline
					\multicolumn{2}{|c|}{} 
					& \tabincell{c}{Upsampled\\flow} 
					& \tabincell{c}{Super-resolved\\flow} 
					& \tabincell{c}{\textcolor{black}{FlowNet-S}\\ \cite{2015-FlowNet:LearningOpticalFlowwithConvolutionalNetworks-Dosovitskiy-2758-2766}}
					& \tabincell{c}{\textcolor{black}{SpyNet}\\ \cite{2017-OpticalFlowEstimationUsingaSpatialPyramidNetwork-Ranjan-2720-2729}}
					\tabularnewline
					\cline{1-6}
					\multicolumn{2}{|c|}{Sintel clean}  
					& 10.96 &  \textbf{10.58} & 7.21 & \textbf{4.63}
					\tabularnewline
					\hline
					\multicolumn{2}{|c|}{Sintel final} 
					& 11.21 &  \textbf{10.83} & 8.17 & \textbf{6.02}
					\tabularnewline
					\hline
					\multicolumn{2}{|c|}{Middlebury} 
					& 1.69 & \textbf{1.30} & 1.18 & \textbf{0.81}
					\tabularnewline
					\hline
					\multirow{2}{*}{KITTI 2012}& Noc 
					& 23.02 & \textbf{22.24} & 13.55 & \textbf{7.62}
					\tabularnewline
					& All & 30.03 & \textbf{29.28} & 19.24 & \textbf{13.30}
					\tabularnewline
					\hline
					\multirow{2}{*}{KITTI 2015}& Noc 
					& 24.64 & \textbf{23.82} & 20.39 & \textbf{15.34}
					\tabularnewline
					& All & 33.27 & \textbf{32.47} & 28.52 & \textbf{23.61} 
					\tabularnewline
					\hline
					
			\end{tabular}}
		\end{center}
	\end{table}
	
	\begin{table}[t]
		\caption{\textcolor{black}{Average RMSE and PSNR results achieved on the Vid4 dataset for $4\times$ SR. Best results are shown in boldface.}}
		\label{tab3}
		\begin{center}
			\normalsize
			\setlength{\tabcolsep}{3mm}{
				\begin{tabular}{|c|c|c|c|c|}
					\hline
					&  \multicolumn{2}{c|}{Upsampled flow} &  \multicolumn{2}{c|}{Super-resolved flow}
					\tabularnewline
					\cline{2-5}
					& \tabincell{c}{RMSE\\($\times10^{-2}$)} & PSNR & \tabincell{c}{RMSE\\($\times10^{-2}$)} & PSNR
					\tabularnewline
					\hline 
					\emph{Calendar} &  4.76 & 26.51 & \textbf{4.56}  &  \textbf{26.89}
					\tabularnewline
					\emph{City} 	& 3.09  & 30.49 & \textbf{3.04}  &  \textbf{30.62}
					\tabularnewline
					\emph{Foliage}  & 3.00  & 30.49 & \textbf{2.71}  &  \textbf{31.37}
					\tabularnewline
					\emph{Walk} 	& 2.99  & 30.56 & \textbf{2.74}  &  \textbf{31.31}
					\tabularnewline
					\hline
					Average  & 3.46  & 29.51  & \textbf{3.26}  & \textbf{30.05}
					\tabularnewline
					\hline
			\end{tabular}}
		\end{center}
	\end{table}
	
	\begin{figure*}[t]
		\centering
		\includegraphics[width=0.82\linewidth]{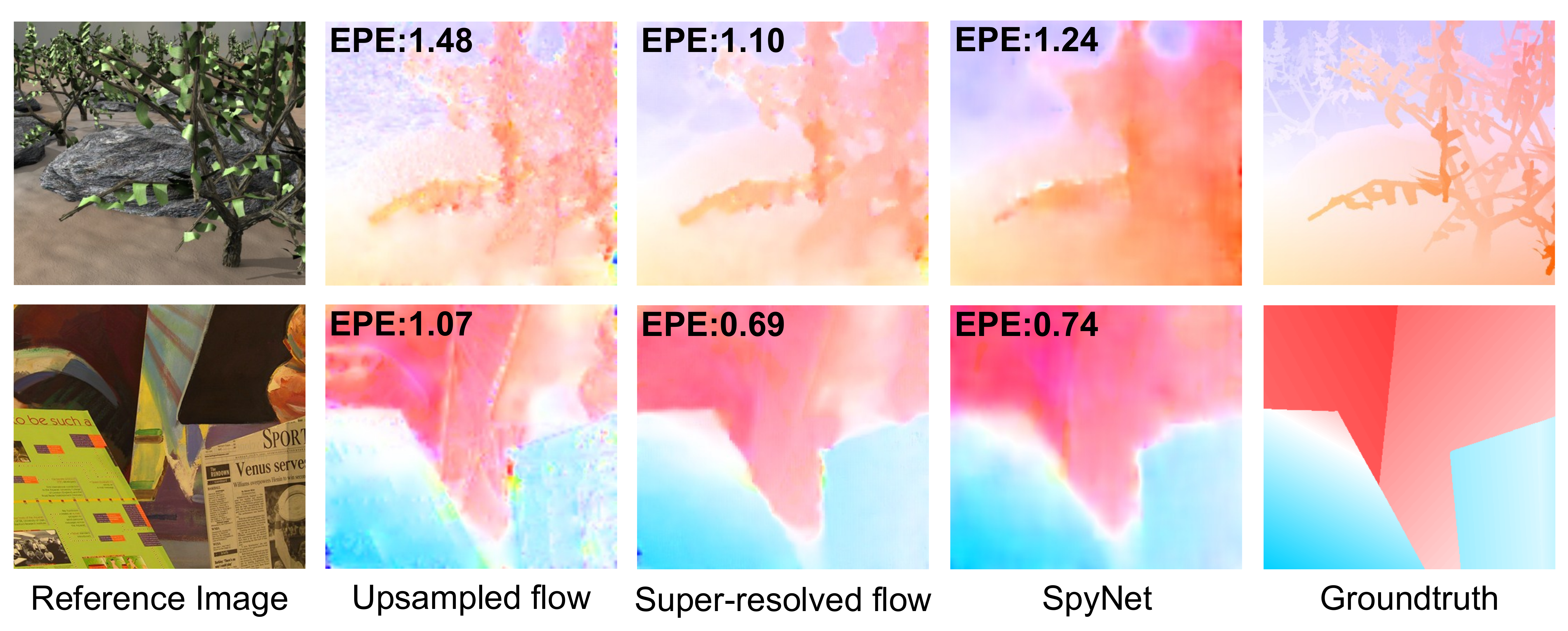}
		\caption{\textcolor{black}{Visual comparison of optical flow estimation results achieved on the Middlebury dataset for $4\times$ SR. The super-resolved optical flows recover finer correspondences with more clear edges and fewer artifacts \textcolor{black}{than the upsampled optical flows}.}}
		\label{fig6}
	\end{figure*}
	\begin{figure}[t]
		\centering
		\includegraphics[width=1\linewidth]{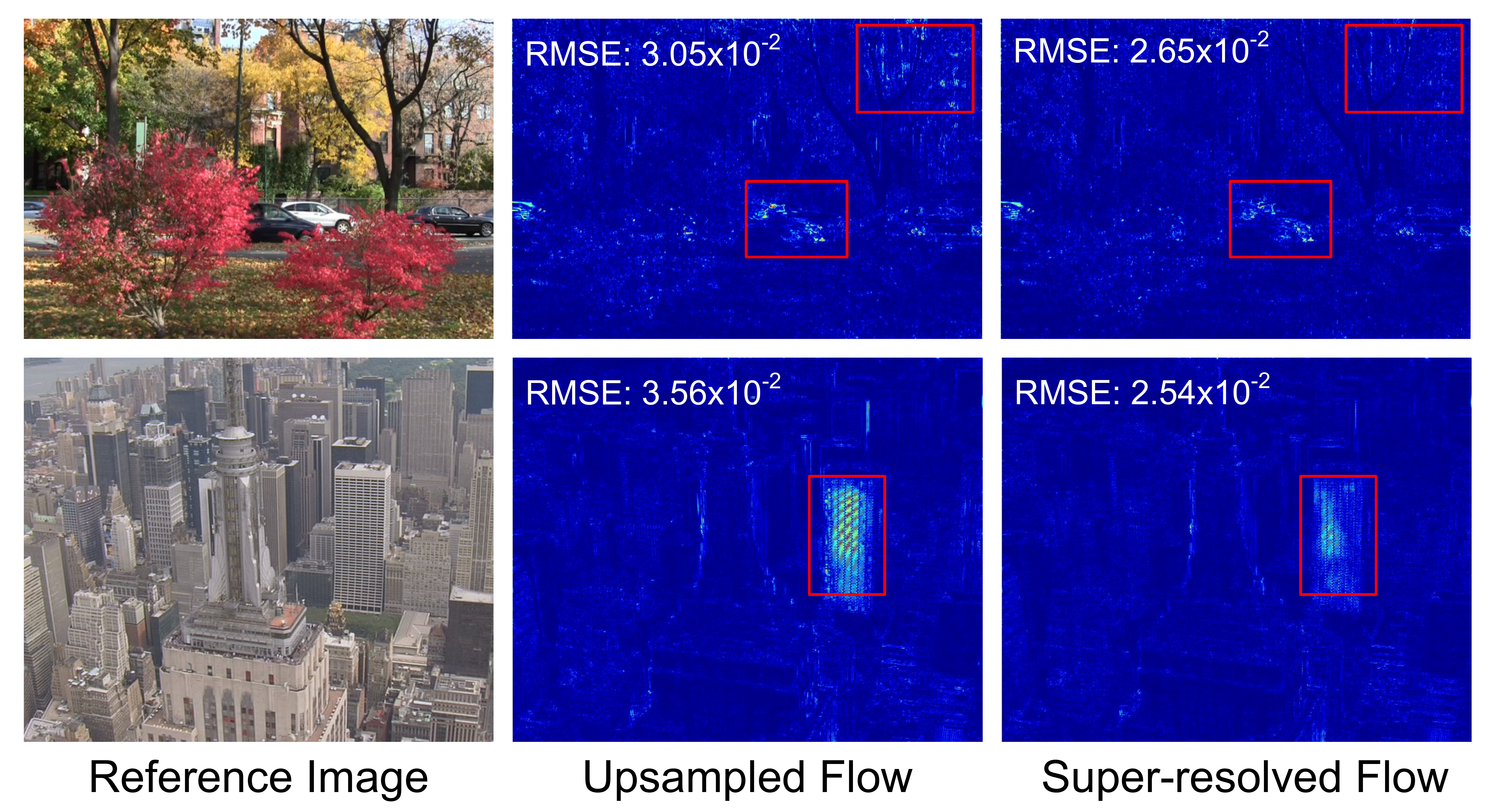}
		\caption{\textcolor{black}{Visual comparison of error maps (difference between the warped image and the reference image) achieved on the Vid4 dataset for $4\times$ SR. The results generated with super-resolved optical flows achieve higher accuracy.}}
		\label{fig7}
	\end{figure}
	
	We further compare the super-resolved optical flows and upsampled optical flows to the groundtruth on the Sintel \cite{2012-ANaturalisticOpenSourceMovieforOpticalFlowEvaluation-Butler-611-625}, Middlebury \cite{2011-ADatabaseandEvaluationMethodologyforOpticalFlow-Baker-1-31}, KITTI 2012 \cite{2012-AreWeReadyforAutonomousDriving?theKITTIVisionBenchmarkSuite-Geiger-3354-3361} and KITTI 2015 \cite{2015-ObjectSceneFlowforAutonomousVehicles-Menze-3061-3070} datasets. \textcolor{black}{We also include two dedicated optical flow estimation methods for comparison, including FlowNet \cite{2015-FlowNet:LearningOpticalFlowwithConvolutionalNetworks-Dosovitskiy-2758-2766} and SpyNet \cite{2017-OpticalFlowEstimationUsingaSpatialPyramidNetwork-Ranjan-2720-2729}. Note that, the optical flows estimated from LR frames are upsampled for fair evaluation.}
	We use the average end-point error (EPE) for quantitative comparison, and present the results in Table \ref{tab2}. 
	
	It can be observed that super-resolved optical flows significantly outperform upsampled ones, with EPE results being reduced by over 0.3. 
	\textcolor{black}{Note that, FlowNet (30.58M) and SpyNet (1.14M) are trained on a much larger dataset (\emph{i.e.} the Flying Chairs dataset with 22872 image pairs) in a supervised manner. Therefore, they achieve better performance than our OFRnet (0.41M) in terms of EPE.}
	Since  groundtruth optical flows are unavailable for the Vid4 dataset, we warped frames using the generated flows and then calculated root mean square error (RMSE) for quantitative evaluation. From Table \ref{tab3} we can also see that images warped using super-resolved optical flows have lower RMSE values (3.26 \emph{vs.} 3.46) and higher PSNR values (30.05 \emph{vs.} 29.51).

	\textcolor{black}{Visual comparison of optical flow estimation results achieved on the Sintel and Middlebury datasets is shown in Figs. \ref{fig5} and \ref{fig6}, respectively. It can be observed that upsampled optical flows produce distorted and blurred edges (\emph{e.g.}, the hand in Fig. \ref{fig5} and the bush in Fig. \ref{fig6})  with notable artifacts. In contrast, more clear edges can be observed in super-resolved optical flows, with finer details being recovered. \textcolor{black}{Moreover, our OFRnet produces visually comparable flow estimation results to SpyNet. This has clearly demonstrated the effectiveness of our OFRnet in recovering temporal details.}
	Error maps achieved on two scenes of the Vid4 dataset are further shown in Fig. \ref{fig7}. It can be observed that super-resolved optical flows produce fewer erroneous pixels, i.e, finer temporal details are recovered. }
	
	\textcolor{black}{In summary, the superior performance achieved on the Sintel, Middlebury, KITTI 2012, KITTI 2015 and Vid4 datasets demonstrates that finer temporal details can be recovered in super-resolved optical flows than upsampled ones. \textcolor{black}{Note that, the task of our work is not to design a superior optical flow estimation network. Instead, we focus on the design of a lightweight sub-network, which is sufficiently effective to provide fine temporal details for the improvement of overall video SR performance.}}
	
	\textcolor{black}{
	\subsubsection{SISR before Optical Flow Estimation.} 
	To obtain HR optical flows from LR inputs, an alternative is to  perform single image super-resolution (SISR) on separated LR frames first and then estimate HR optical flows from these SR results. To test the performance of this option, we designed a variant to perform SISR before optical flow estimation. Specifically, input LR frames were first super-resolved separately before being fed to the OFRnet for HR optical flow estimation. Note that, the sub-pixel convolution in level 3 of the OFRnet was replaced with a normal convolution. Next, SISR results were compensated and passed to the SRnet for fusion.}
	\textcolor{black}{It can be observed from Table \ref{tab1} that this variant does not introduce significant performance improvement against our SOF-VSR in terms of PSNR and SSIM. Meanwhile, this variant requires much higher computational cost than our SOF-VSR, with FLOPs being increased from 108.90G to 1.12T. Since SISR is first used to enhance the resolution of LR inputs, optical flow estimation and fusion of multiple frames are performed on HR images. Therefore, the computational complexity is significantly increased. In contrast, our SOF-VSR directly infers HR optical flows  from LR inputs and fuses multiple frames in LR space. Therefore, our network has a much lower computational cost and is more suitable for applications on mobile computing devices.} 
		
	\textcolor{black}{		
	\subsubsection{Scale-recurrent vs. Scale-cascaded Architectures} 
	Since the task of each level in our OFRnet is similar, we employ a scale-recurrent architecture in our OFRnet to reduce model size. To demonstrate its effectiveness, we replaced the scale-recurrent architecture with a scale-cascaded one by using independent networks at 3 levels. Results achieved on the Vid4 dataset are presented in Table \ref{tab1}.}
	
	\textcolor{black}{
	Our SOF-VSR achieves comparable performance to the scale-cascaded architecture with the overall model size being reduced from 1.33M to 1.00M. Since the tasks of different levels are similar, the scale-cascaded architecture contains redundant parameters. In contrast, using a scale-recurrent structure, our SOF-VSR is more  lightweight and compact while achieving comparable performance.}

	\begin{table*}[ht]
		\caption{Comparative results achieved on the Vid4 dataset. Note that, the first and last two frames are not used in our evaluation. FLOPs is computed based on HR frames with a resolution of 720p (1280$\times$720). Results marked with * are directly copied from the corresponding papers. Best results are shown in boldface.}
		\label{tab4}
		\begin{center}
			\normalsize
			\setlength{\tabcolsep}{2mm}{
				\begin{tabular}{|c|c|l|c|c|c|c|c|c|c|}
					\hline 
					Model & Scale & Method  & Frames & PSNR($\uparrow$) & SSIM($\uparrow$) & \tabincell{c}{T-MOVIE($\downarrow$)\\($\times10^{-3}$)}  &\tabincell{c}{MOVIE($\downarrow$)\\($\times10^{-3}$)} & \textcolor{black}{Params.} & \textcolor{black}{FLOPs}
					\tabularnewline
					\hline
					\hline
					\multirow{22}{*}{BI} & \multirow{6}{*}{$\times2$} & Bicubic & 1 & 28.42 & 0.866 & 7.24 & 1.35  & - & -
					\tabularnewline				
					& & DRCN \cite{2016-DeeplyRecursiveConvolutionalNetworkforImageSuperResolution-Kim-1637-1645}  & 1 & 31.57 & 0.924 & 2.71 & 0.46 & 1.8M & 9,788.7G
					\tabularnewline
					& & LapSRN \cite{2017-DeepLaplacianPyramidNetworksforFastandAccurateSuperResolution-Lai-5835-5843} & 1 & 31.41 & 0.923 &  2.62 & 0.45 & 0.8M & 29.9G
					\tabularnewline
					& & CARN \cite{2018-FastAccurateandLightweightSuperResolutionwithCascadingResidualNetwork-Ahn--}  & 1 & 31.96 & 0.931 & 2.35 & 0.39 & 1.6M & 222.8G		
					\tabularnewline
					& & VSRnet \cite{2016-VideoSuperResolutionwithConvolutionalNeuralNetworks-Kappeler-109-122} & 5 & 31.29 & 0.927 & 2.93 & 0.48  & 266K & 242.7G
					\tabularnewline					& & SOF-VSR  & 3 & \textbf{33.17} & \textbf{0.947} & \textbf{1.43} & \textbf{0.23} & 0.9M & 342.8G
					\tabularnewline
					\cline{2-10}
					& \multirow{6}{*}{$\times3$} & Bicubic  & 1 & 25.26 & 0.730 & 21.15 &  5.02 & - & -
					\tabularnewline
					& & DRCN \cite{2016-DeeplyRecursiveConvolutionalNetworkforImageSuperResolution-Kim-1637-1645} & 1 & 26.82 & 0.805 & 12.15 & 3.03 & 1.8M & 9,788.7G
					\tabularnewline
					& & CARN \cite{2018-FastAccurateandLightweightSuperResolutionwithCascadingResidualNetwork-Ahn--} & 1 & 27.16 & 0.818 & 10.69 & 2.65 & 1.6M & 118.8G
					\tabularnewline
					& & VSRnet \cite{2016-VideoSuperResolutionwithConvolutionalNeuralNetworks-Kappeler-109-122}  & 5 & 26.75 & 0.807 & 12.14 & 2.81 & 266K & 242.7G
					\tabularnewline
					& & VESCPN \cite{2017-RealTimeVideoSuperResolutionwithSpatioTemporalNetworksandMotionCompensation-Caballero-2848-2857} & 3 & 27.25* & 0.845* & - & 2.86* & 89K & 5.3G
					\tabularnewline
					& & SOF-VSR & 3 & \textbf{28.09} & \textbf{0.861} & \textbf{8.25} & \textbf{1.83}  & 1.1M &205.0G
					\tabularnewline
					\cline{2-10}
					& \multirow{8}{*}{$\times4$} & Bicubic  & 1 & 23.75 & 0.630 & 35.93 & 8.80 & - & -
					\tabularnewline
					& & DRCN \cite{2016-DeeplyRecursiveConvolutionalNetworkforImageSuperResolution-Kim-1637-1645}  & 1 & 24.94 & 0.707 & 25.48 & 6.28 & 1.8M & 9,788.7G
					\tabularnewline
					& & LapSRN \cite{2017-DeepLaplacianPyramidNetworksforFastandAccurateSuperResolution-Lai-5835-5843}  & 1 & 24.98 & 0.711 & 24.93 & 6.05 & 0.8M & 149.4G
					\tabularnewline 
					& & CARN \cite{2018-FastAccurateandLightweightSuperResolutionwithCascadingResidualNetwork-Ahn--} & 1 & 25.27 & 0.725 & 21.95 & 5.59 & 1.6M & 90.9G
					\tabularnewline
					& & VSRnet \cite{2016-VideoSuperResolutionwithConvolutionalNeuralNetworks-Kappeler-109-122} & 5 & 24.81 & 0.702 & 26.05 & 6.01 & 266K & 242.7G
					\tabularnewline
					& & VESCPN \cite{2017-RealTimeVideoSuperResolutionwithSpatioTemporalNetworksandMotionCompensation-Caballero-2848-2857} & 3 &25.35* & 0.756* & - & 5.82* & 91K & 3.1G
					\tabularnewline
					& & TDVSR \cite{2017-RobustVideoSuperResolutionwithLearnedTemporalDynamics-Liu--} & 5 & 25.49 & 0.746 & 23.23 & 4.92 & 343K & 24.7G
					\tabularnewline
					& & TDVSR-L \cite{2018-LearningTemporalDynamicsforVideoSuperResolution:aDeepLearningApproach-Liu--} & 5 & 25.88* & 0.767* & - & 4.69* & 2.0M & 263.9G
					\tabularnewline
					& & SOF-VSR \cite{2018-LearningforVideoSuperResolutionthroughHROpticalFlowEstimation-LongguangWang--}  & 3 & \textbf{26.01}* & 0.771* & 19.78* & 4.32* & 1.5M & 105.2G
					\tabularnewline
					& & SOF-VSR & 3 & 26.00 &\textbf{ 0.772} & \textbf{19.35} & \textbf{4.25} & 1.0M & 112.5G
					\tabularnewline
					\hline
					\hline
					\multirow{4}{*}{BD} & \multirow{4}{*}{$\times4$}& SPMC \cite{2017-DetailRevealingDeepVideoSuperResolution-Tao-4482-4490} & 3 & 25.99 & 0.773 & 18.28 & 4.00 & 1.7M & 160.8G
					\tabularnewline
					& & FRVSR-3-64 \cite{2018-FrameRecurrentVideoSuperResolution-Sajjadi-6626-6634} & Recurrent & 26.17* & \textbf{0.798*} & - & -  & 2.3M & 88.8G
					\tabularnewline
					& & SOF-VSR-BD \cite{2018-LearningforVideoSuperResolutionthroughHROpticalFlowEstimation-LongguangWang--}  & 3 & \textbf{26.19}* & 0.785* & 17.63* & 4.00*  & 1.5M & 105.2G
					\tabularnewline
					& & SOF-VSR-BD & 3 & \textbf{26.19} & 0.786 &\textbf{17.61} & \textbf{3.98}  & 1.0M & 112.5G
					\tabularnewline
					\hline
			\end{tabular}}
		\end{center}
	\end{table*}

	\textcolor{black}{
	\subsubsection{Efficient Residual Block vs. Vanilla Residual Block} 
	Efficient residual block is used in our SOF-VSR network to reduce  model size and computational complexity. To demonstrate its effectiveness, we designed a variant by replacing efficient residual blocks with vanilla ones. Comparative results are listed in Table \ref{tab1}.}
	
	\textcolor{black}{
	It can be observed that the variant with vanilla residual blocks achieves slightly better performance than our SOF-VSR. However, its model size and FLOPs are increased from 1.00M to 1.56M and from 108.90G to 143.14G, respectively. Using efficient residual blocks, our SOF-VSR is more lightweight and compact  without obvious performance drop. Therefore, our SOF-VSR network is more suitable for applications on mobile computing devices.}

	\subsection{Comparison to the State-of-the-Art}
	
	We compared our SOF-VSR to 4 single image SR methods including Bicubic, Deeply Recursive Convolutional Network (DRCN) \cite{2016-DeeplyRecursiveConvolutionalNetworkforImageSuperResolution-Kim-1637-1645}, Laplacian Pyramid Super-Resolution Network (LapSRN) \cite{2017-DeepLaplacianPyramidNetworksforFastandAccurateSuperResolution-Lai-5835-5843}, and Cascading Residual Network (CARN) \cite{2018-FastAccurateandLightweightSuperResolutionwithCascadingResidualNetwork-Ahn--} and 6 video SR methods including Video Super-Resolution Network (VSRnet) \cite{2016-VideoSuperResolutionwithConvolutionalNeuralNetworks-Kappeler-109-122},
	VESCPN \cite{2017-RealTimeVideoSuperResolutionwithSpatioTemporalNetworksandMotionCompensation-Caballero-2848-2857},
	TDVSR \cite{2017-RobustVideoSuperResolutionwithLearnedTemporalDynamics-Liu--},
	TDVSR-L \cite{2018-LearningTemporalDynamicsforVideoSuperResolution:aDeepLearningApproach-Liu--},
	SPMC \cite{2017-DetailRevealingDeepVideoSuperResolution-Tao-4482-4490},
	and FRVSR \cite{2018-FrameRecurrentVideoSuperResolution-Sajjadi-6626-6634} 
	on the Vid4 and DAVIS-10 datasets. For DRCN, LapSRN, CARN, VSRnet, and SPMC, we used the codes provided by the authors to produce their results. For TDVSR, we used the super-resolved images provided by the authors. For VESCPN, TDVSR-L and FRVSR, the results reported in their papers are used.
	Here, we only report the performance of FRVSR-3-64 since its network size are comparable to ours. For each test video clip with 31 frames, the first and last two frames are not used for performance evaluation.
	
	Note that, the methods selected for comparison are trained on two different degradation models. Specifically, the degradation model used in DRCN, LapSRN, CARN,  VSRnet, VESCPN,  TDVSR and TDVSR-L is bicubic downsampling (denoted as BI), which is implemented using Matlab function $imresize$. For SPMC and FRVSR, HR images are first blurred using a Gaussian kernel and then downsampled by selecting every $s^{\textup{th}}$ pixel (denoted as BD). Consequently, we retrained our SOF-VSR network on the BD degradation model (denoted as SOF-VSR-BD) to achieve fair comparison with SPMC and FRVSR. In this work, a Gaussian kernel with a standard deviation $\sigma\!=\!1.6$ is used, which is the same as the one used in SPMC, but slightly larger than the one used in FRVSR ($\sigma\!=\!1.5$).

	\subsubsection{Evaluation on the Vid4 Dataset}  \indent
	
	\textbf{Quantitative Evaluation.} Quantitative results achieved on the Vid4 dataset are shown in Table \ref{tab4}. 
	For the BI degradation model, our SOF-VSR achieves the best performance for $2\times$ and $3\times$ SR. \textcolor{black}{For $4\times$ SR, our SOF-VSR outperforms TDVSR-L in terms of PSNR, SSIM and MOVIE with halved parameters and FLOPs. Compared to the conference version, our SOF-VSR achieves comparable performance with much fewer parameters (1.0M \emph{vs.} 1.5M). Moreover, our network outperforms other methods in terms of T-MOVIE and MOVIE. That means our results are temporally more consistent.} That is because, more accurate temporal dependency details can be provided by HR optical flows and therefore improved accuracy and consistency performance can be achieved. 
	
	For the BD degradation model, our SOF-VSR-BD network outperforms SPMC, with PSNR, SSIM and T-MOVIE values being improved by a notable margin. Although FRVSR-3-64 achieves a higher SSIM value, our SOF-VSR-BD method still achieves comparable performance in terms of other metrics with halved parameters. Compared to our conference version, our SOF-VSR-BD achieves comparable performance with parameters being reduced by over 30\%.
	
	\begin{figure}[tp]
		\centering
		\includegraphics[width=0.85\linewidth]{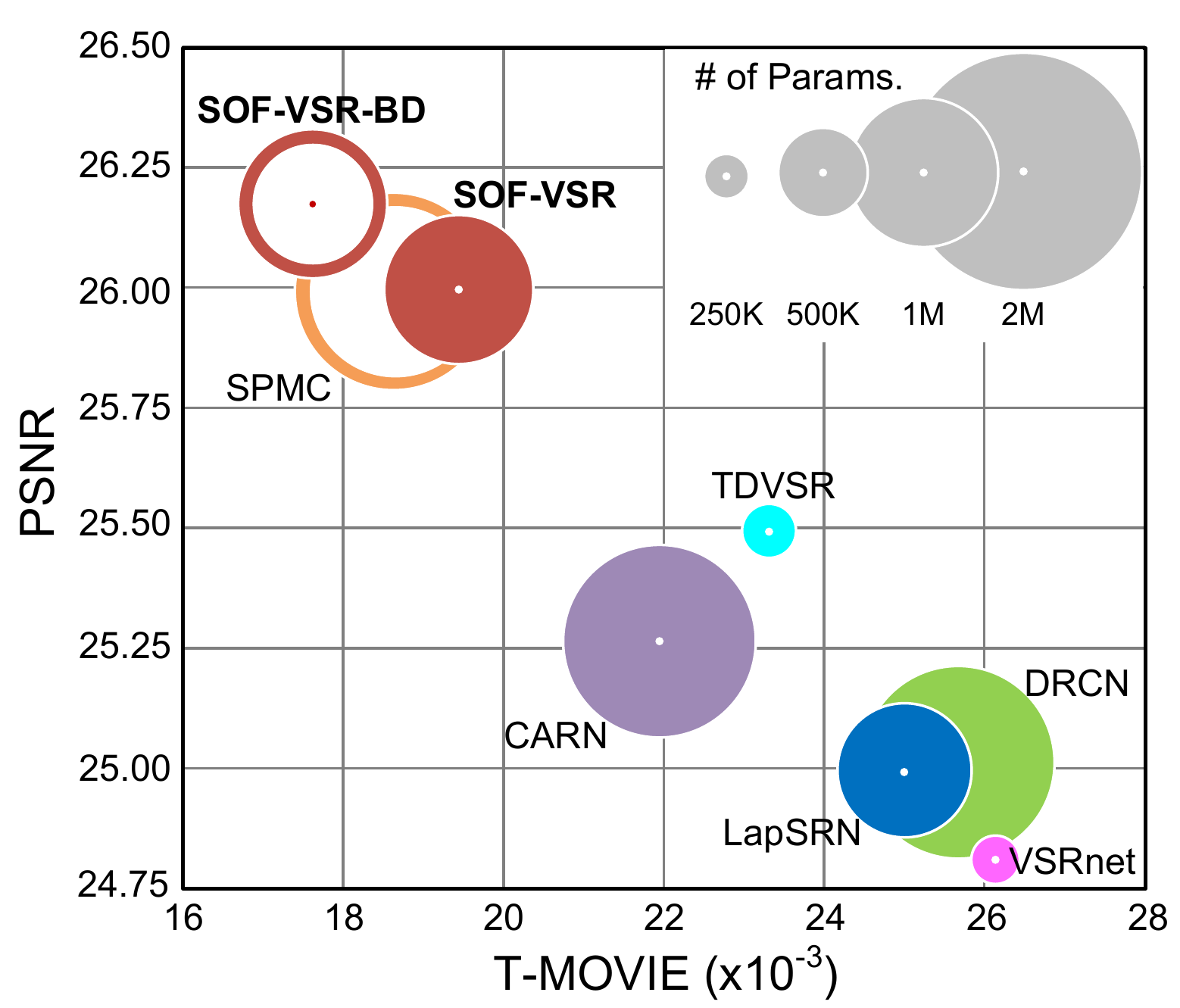}
		\caption{\textcolor{black}{Consistency and accuracy performance achieved on the Vid4 dataset for $4\times$ SR. Solid and hollow circles represent methods developed for the BI and BD degradation models, respectively. A lower T-MOVIE value represents a better consistency performance, while a higher PSNR value represents a better accuracy performance. The size of a circle represents the number of parameters.}}
		\label{fig8}
	\end{figure}
		
	\begin{figure*}[ht]
		\centering
		\includegraphics[width=1\linewidth]{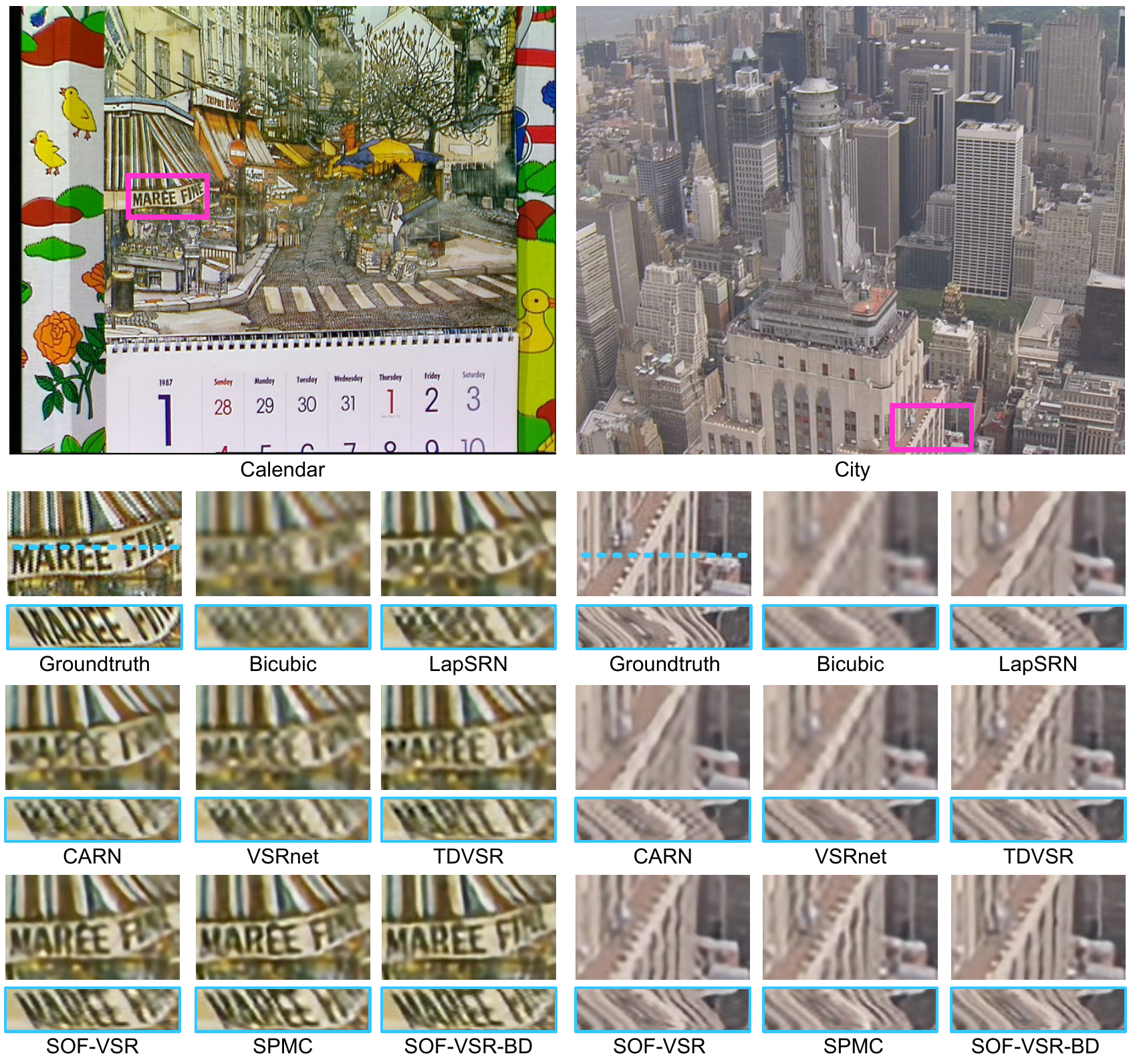}
		\caption{Visual comparison of $4\times$ SR results on \emph{Calendar} and \emph{City}. Bicubic, LapSRN \cite{2017-DeepLaplacianPyramidNetworksforFastandAccurateSuperResolution-Lai-5835-5843}, CARN \cite{2018-FastAccurateandLightweightSuperResolutionwithCascadingResidualNetwork-Ahn--}, VSRnet \cite{2016-VideoSuperResolutionwithConvolutionalNeuralNetworks-Kappeler-109-122}, TDVSR \cite{2017-RobustVideoSuperResolutionwithLearnedTemporalDynamics-Liu--}, and SOF-VSR are based on the BI degradation model, while SPMC \cite{2017-DetailRevealingDeepVideoSuperResolution-Tao-4482-4490} and SOF-VSR-BD are based on the BD degradation model. \textcolor{black}{Blue boxes represent corresponding temporal profiles.}}
		\label{fig9}
	\end{figure*}
	
	\begin{table*}[t]
		\caption{Comparison of accuracy and consistency performance achieved on the DAVIS-10 dataset. Note that, the first and last two frames are not used in our evaluation. FLOPs is computed based on HR frames with a resolution of 720p (1280$\times$720). Best results are shown in boldface.}
		\label{tab5}
		\begin{center}
			\normalsize
			\setlength{\tabcolsep}{2mm}{
				\begin{tabular}{|c|c|l|c|c|c|c|c|c|c|}
					\hline 
					Model & Scale & Method & Frames & PSNR($\uparrow$) & SSIM($\uparrow$) & \tabincell{c}{T-MOVIE($\downarrow$)\\($\times10^{-3}$)}  &\tabincell{c}{MOVIE($\downarrow$)\\($\times10^{-3}$)} & \textcolor{black}{Params.} & \textcolor{black}{FLOPs}
					\tabularnewline
					\hline
					\hline
					\multirow{18}{*}{BI} & \multirow{5}{*}{$\times2$} & Bicubic  & 1 & 36.43 & 0.958 & 4.63 & 0.70 & - & -
					\tabularnewline				
					& & DRCN \cite{2016-DeeplyRecursiveConvolutionalNetworkforImageSuperResolution-Kim-1637-1645} & 1 & 40.62 & 0.979 & 1.09 & 0.13 &1.8M & 9,788.7G
					\tabularnewline
					& & LapSRN \cite{2017-DeepLaplacianPyramidNetworksforFastandAccurateSuperResolution-Lai-5835-5843}  & 1 & 40.30 & 0.978 & 1.05 & 0.12 & 0.8M & 29.9G
					\tabularnewline
					& & CARN \cite{2018-FastAccurateandLightweightSuperResolutionwithCascadingResidualNetwork-Ahn--}& 1 & 40.99 & 0.981 & \textbf{0.85} & 0.11 & 1.6M & 222.8G
					\tabularnewline
					& & VSRnet \cite{2016-VideoSuperResolutionwithConvolutionalNeuralNetworks-Kappeler-109-122} & 5 & 39.00 & 0.972 & 1.31 & 0.20  &  266K & 242.7G
					\tabularnewline
					& & SOF-VSR  & 3 &\textbf{41.38} & \textbf{0.983} & 0.92 & \textbf{0.09} & 0.9M & 342.8G
					\tabularnewline
					\cline{2-10}
					& \multirow{5}{*}{$\times3$} & Bicubic & 1 & 32.94 & 0.912 & 13.55 & 2.63  & - & -
					\tabularnewline
					& & DRCN \cite{2016-DeeplyRecursiveConvolutionalNetworkforImageSuperResolution-Kim-1637-1645}  & 1 & 36.08 & 0.947 & 5.26 & 0.92 & 1.8M & 9,788.7G
					\tabularnewline
					& & CARN \cite{2018-FastAccurateandLightweightSuperResolutionwithCascadingResidualNetwork-Ahn--}  & 1 & 36.70 & 0.952 & 4.44 & 0.79 & 1.6M & 118.8G
					\tabularnewline
					& & VSRnet \cite{2016-VideoSuperResolutionwithConvolutionalNeuralNetworks-Kappeler-109-122}  & 5 & 34.94 & 0.936 & 6.11 & 1.20 &266K & 242.7G
					\tabularnewline					
					& & SOF-VSR  & 3 & \textbf{36.80} & \textbf{0.955} & \textbf{4.36} & \textbf{0.68}  & 1.1M & 205.0G
					\tabularnewline
					\cline{2-10}
					& \multirow{5}{*}{$\times4$} & Bicubic  & 1 & 30.97 & 0.870 & 22.73 & 4.75 & - & -
					\tabularnewline
					& & DRCN \cite{2016-DeeplyRecursiveConvolutionalNetworkforImageSuperResolution-Kim-1637-1645}  & 1 & 33.49 & 0.911 & 13.51 & 2.48 & 1.8M & 9,788.7G
					\tabularnewline
					& & LapSRN \cite{2017-DeepLaplacianPyramidNetworksforFastandAccurateSuperResolution-Lai-5835-5843}  & 1 & 33.54 & 0.911 & 12.83 & 2.43 & 0.8M & 149.4G
					\tabularnewline
					& & CARN \cite{2018-FastAccurateandLightweightSuperResolutionwithCascadingResidualNetwork-Ahn--} & 1 & 34.12 & 0.921 & \textbf{11.41} & 2.05 & 1.6M & 90.9G
					\tabularnewline
					& & VSRnet \cite{2016-VideoSuperResolutionwithConvolutionalNeuralNetworks-Kappeler-109-122}  & 5 & 32.63 & 0.897 & 14.63 & 2.85 & 266K & 242.7G
					\tabularnewline
					& & SOF-VSR \cite{2018-LearningforVideoSuperResolutionthroughHROpticalFlowEstimation-LongguangWang--}  & 3 & \textbf{34.32}* & 0.925* & 11.77* & 1.96* & 1.5M  & 105.2G
					\tabularnewline
					& & SOF-VSR  & 3 & 34.28 & \textbf{0.926} & 11.72 & \textbf{1.94} & 1.0M & 112.5G
					\tabularnewline
					\hline
					\hline
					\multirow{3}{*}{BD} & \multirow{3}{*}{$\times4$}& SPMC \cite{2017-DetailRevealingDeepVideoSuperResolution-Tao-4482-4490}  & 3 & 33.02 & 0.911 & 14.06 & 1.96  &1.7M & 160.8G 
					\tabularnewline
					& & SOF-VSR-BD \cite{2018-LearningforVideoSuperResolutionthroughHROpticalFlowEstimation-LongguangWang--}  & 3 & 34.27* & 0.925* & 10.93* & 1.90* & 1.5M & 105.2G
					\tabularnewline
					& & SOF-VSR-BD  & 3 & \textbf{34.28} & \textbf{0.927} & \textbf{10.91} & \textbf{1.87}  & 1.0M & 112.5G
					\tabularnewline
					\hline
			\end{tabular}}
		\end{center}
	\end{table*}
	
	\begin{figure*}[t]
		\centering
		\includegraphics[width=1\linewidth]{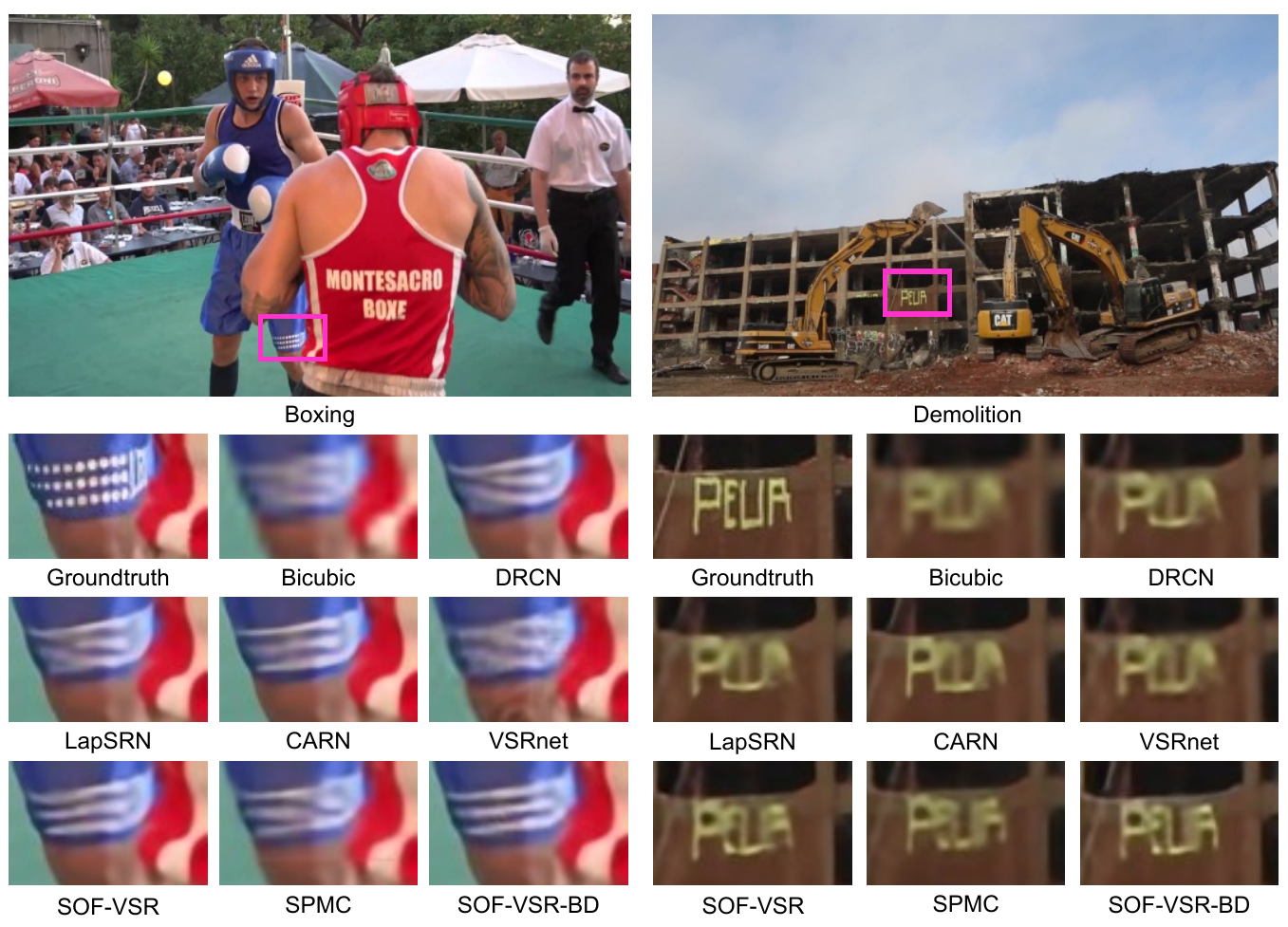}
		\caption{Visual comparison of $4\times$ SR results on \emph{Boxing} and \emph{Demolition}. Bicubic, DRCN \cite{2016-DeeplyRecursiveConvolutionalNetworkforImageSuperResolution-Kim-1637-1645}, LapSRN \cite{2017-DeepLaplacianPyramidNetworksforFastandAccurateSuperResolution-Lai-5835-5843}, CARN \cite{2018-FastAccurateandLightweightSuperResolutionwithCascadingResidualNetwork-Ahn--}, VSRnet \cite{2016-VideoSuperResolutionwithConvolutionalNeuralNetworks-Kappeler-109-122}, and SOF-VSR are based on the BI degradation model, while SPMC \cite{2017-DetailRevealingDeepVideoSuperResolution-Tao-4482-4490} and SOF-VSR-BD are based on the BD degradation model.}
		\label{fig10}
	\end{figure*}
	
	We further show the trade-off between accuracy and consistency of different methods in Fig. \ref{fig8}. It can be observed that our SOF-VSR and SOF-VSR-BD networks achieve better PSNR and T-MOVIE performance on the Vid4 dataset, while being lightweight and compact.
	
	\textbf{Qualitative Evaluation.}
	Several qualitative results on two scenarios of the Vid4 dataset are shown in Fig. \ref{fig9}. We can see from the zoom-in regions that our SOF-VSR and SOF-VSR-BD networks recover finer details, such as the word ``MAREE" and the stripes of the building. Moreover, it can be observed from the temporal profiles that the word ``MAREE" can hardly be recognized in the SR results achieved by Bicubic, DRCN, LapSRN, CARN, VSRnet and TDVSR. Although finer results are produced by SPMC, the word is still distorted and blurred. In contrast, smooth and clear patterns with fewer artifacts can be observed in the temporal profiles of our results. In summary, our network produces temporally more consistent results and better perceptual quality.

	\subsubsection{Evaluation on the DAVIS-10 Dataset} \indent
	
	\textbf{Quantitative Evaluation.} 
	Quantitative results achieved on the DAVIS-10 dataset are shown in Table \ref{tab5}. For the BI degradation model, our SOF-VSR network achieves the state-of-the-art performance in terms of PSNR and SSIM. Although suffering from a slight PSNR performance drop as compared to the conference version for $4\times$ SR, our network  achieves better performance in terms of other metrics with much fewer parameters (1.0M \emph{vs.} 1.5M). In terms of T-MOVIE, our network achieves comparable or better performance than other approaches. In summary, our SOF-VSR network produces SR results with the best overall video quality in terms of MOVIE. 
	
	For the BD degradation model, our SOF-VSR-BD network outperforms SPMC in terms of all metrics. Specifically, the PSNR/T-MOVIE values achieved by our network are better than SPMC by \textcolor{black}{1.26}/3.15. That is, better accuracy and consistency performance is achieved by our network. Since the DAVIS-10 dataset comprises scenes with fast moving objects, complex motion patterns (especially large displacements) lead to performance deterioration of existing video SR methods. In contrast, more accurate temporal dependency is provided by HR optical flows in our network. Therefore, complex motion patterns can be handled more robustly and better performance can be achieved.
	
	\textbf{Qualitative Evaluation.} Qualitative comparison on two scenarios of the DAVIS-10 dataset is shown in Fig. \ref{fig10}. Compared to other methods, our SOF-VSR and SOF-VSR-BD networks recover more accurate details and achieve better visual quality, such as the pattern on the shorts and the word ``PEUA". Specifically, the patterns on the shorts recovered by Bicubic and VSRnet are obviously blurred. Although finer details can be recovered by DRCN, LapSRN and SPMC, their resulting patterns are still distorted. In contrast, our networks produce more clear details with fewer artifacts.

	\subsection{High-Level Vision Tasks}

	\textcolor{black}{
	Rich details in a video clip are beneficial to high-level vision tasks such as face recognition and digit recognition \cite{2016-IsImageSuperResolutionHelpfulforOtherVisionTasks?-Dai-1-9}. Here, we further compare our network to LapSRN, CARN, and SPMC by integrating a video SR module into the face recognition task.}
	
	\textcolor{black}{
	\textbf{Data preparation.} Following  \cite{2018-LearningTemporalDynamicsforVideoSuperResolution:aDeepLearningApproach-Liu--}, we form a subset of the YouTube Face dataset \cite{2011-FaceRecognitioninUnconstrainedVideoswithMatchedBackgroundSimilarity-Wolf-529-534} by choosing 167 subject classes that contain more than three video sequences. For each class, we randomly select one video for test and the rest for training. We first cropped face regions and resized them to the size of 60$\times$60 to generate the HR data. Then, these HR data were downsampled to 15$\times$15 to form the LR data. For each test video, we splitted it into clips of 50 frames. In total, we have about 600 clips.}
	
	\textcolor{black}{
	\textbf{Classifier.} We used a customized AlexNet in \cite{2012-ImagenetClassificationwithDeepConvolutionalNeuralNetworks-Krizhevsky-1097-1105} as the classifier, whose architecture details are shown in Table \ref{tab6}. The classification network takes a 60$\times$60 facial image as input and predicts the class of the subject.} 
	
	\begin{table}[t]
		\caption{\textcolor{black}{Network architecture of the classifier used for face recognition.}}
		\label{tab6}
		\begin{center}
			\normalsize
			\setlength{\tabcolsep}{2mm}{
				\begin{tabular}{|l|c|c|c|}
					\hline  
					Layer Settings & Output Size
					\tabularnewline
					\hline
					Input & $1\times60\times60$
					\tabularnewline
					9$\times$9 Conv, stride 1, padding 0, ReLU & $64\times52\times52$
					\tabularnewline
					5$\times$5 Conv, stride 1, padding 0, ReLU & $32\times48\times48$
					\tabularnewline
					4$\times$4 Conv, stride 1, padding 0, ReLU & $60\times45\times45$
					\tabularnewline 
					Max pooling, kernel 2, stride 2, padding 0 & $60\times23\times23$
					\tabularnewline
					3$\times$3 Conv, stride 1, padding 0, ReLU & $80\times21\times21$
					\tabularnewline
					Max pooling, kernel 2, stride 2, padding 0 & $80\times11\times11$
					\tabularnewline
					Fully connected & 167
					\tabularnewline
					\hline
			\end{tabular}}
		\end{center}
	\end{table}
	
	\begin{table}[t]
		\caption{\textcolor{black}{Face recognition performance achieved by different SR methods on a subset of the YouTube Face dataset under $4\times$ SR scenario. Best results are shown in boldface.}}
		\label{tab7}
		\begin{center}
			\normalsize
			\setlength{\tabcolsep}{1.5mm}{
				\begin{tabular}{|l|c|c|c|c|c|}
					\hline 
					Model & Model  & Top-1 Accuracy & Top-5 Accuracy  
					\tabularnewline
					\hline
					Original    & -  & 62.7\%  & 75.1\%     
					\tabularnewline
					\hline
					Bicubic    &  \multirow{4}{*}{BI} & 41.1\%  & 58.8\%     
					\tabularnewline
					LapSRN \cite{2017-DeepLaplacianPyramidNetworksforFastandAccurateSuperResolution-Lai-5835-5843}
					&   & 55.9\% & 71.4\%	   
					\tabularnewline
					CARN \cite{2018-FastAccurateandLightweightSuperResolutionwithCascadingResidualNetwork-Ahn--} 	
					& & 56.9\% & 72.4\%   
					\tabularnewline
					SOF-VSR &   & \textbf{58.1\%}  & \textbf{74.1\%}     
					\tabularnewline
					\hline
					SPMC \cite{2017-DetailRevealingDeepVideoSuperResolution-Tao-4482-4490}       
					&  \multirow{2}{*}{BD}  &  55.8\% & 71.4\%     
					\tabularnewline
					SOF-VSR-BD  
					&   & \textbf{57.6\%} & \textbf{72.9\%}     
					\tabularnewline
					\hline
			\end{tabular}}
		\end{center}
	\end{table}

	\textcolor{black}{
	\textbf{Implementation details.} During test, we first super-resolved each LR test clip using a specific SR method and then fed the SR results to the classifier. Note that, we did not fine-tune these SR methods on the Youtube Face dataset. The prediction probabilities were aggregated over all frames in each video clip. The top-1 and top-5 accuracy metrics were used for quantitative evaluation and the comparative results are shown in Table \ref{tab7}.}
	
	\textcolor{black}{
	It can be observed that our SOF-VSR network achieves the highest top-1 and top-5 accuracy on both BI and BD degradation models. Specifically, our network outperforms CARN by 1.2\%/1.7\% in terms of top-1/top-5 accuracy. That is because, our SOF-VSR network can recover richer details such that better face recognition performance can be achieved.}

	\section{Conclusion}
	In this paper, we have proposed an end-to-end deep network for video SR. Our OFRnet first super-resolves optical flows to provide accurate temporal dependency. Motion compensation is then performed based on HR optical flows. Finally, SRnet is used to infer SR results from these compensated LR frames. Extensive experimental results show that our SOF-VSR network can recover accurate temporal details for the improvement of both SR accuracy and consistency. Comparison to existing video SR methods has also demonstrated the state-of-the-art performance of our SOF-VSR network.

	\section*{Acknowledgment}
	This work was partially supported by the National Natural Science Foundation of China (No. 61972435, 61602499 and No. 61605242), and Fundamental Research Funds for the Central Universities (No. 18lgzd06).

	\ifCLASSOPTIONcaptionsoff
	\newpage
	\fi



	
	

	\bibliographystyle{IEEEtran}
	\bibliography{IEEEabrv,super-resolution,other-CV-fields,neural-network,image-deblur}
	
	\begin{IEEEbiography}[{\includegraphics[width=1in,height=1.25in,clip]{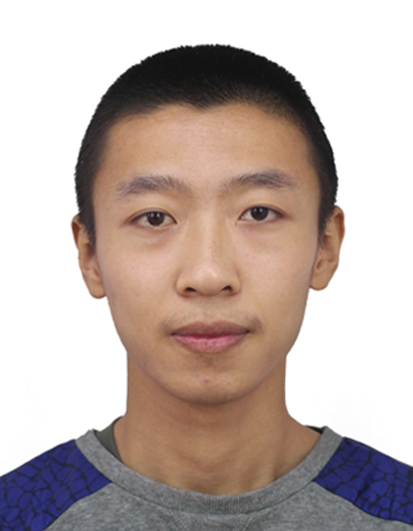}}]
		{Longguang Wang} received the B.E. degree in electric engineering from Shandong University (SDU), Jinan, China, in 2015, and the M.E. degree in information and communication engineering from National University of Defense Technology (NUDT), Changsha, China, in 2017. He is currently pursuing the Ph.D. degree with the College of Electronic Science and Technology, NUDT. His current research interests include low-level vision and deep learning.
	\end{IEEEbiography}
	\vspace{-5ex}
	\begin{IEEEbiography}[{\includegraphics[width=1in,height=1.25in,clip]{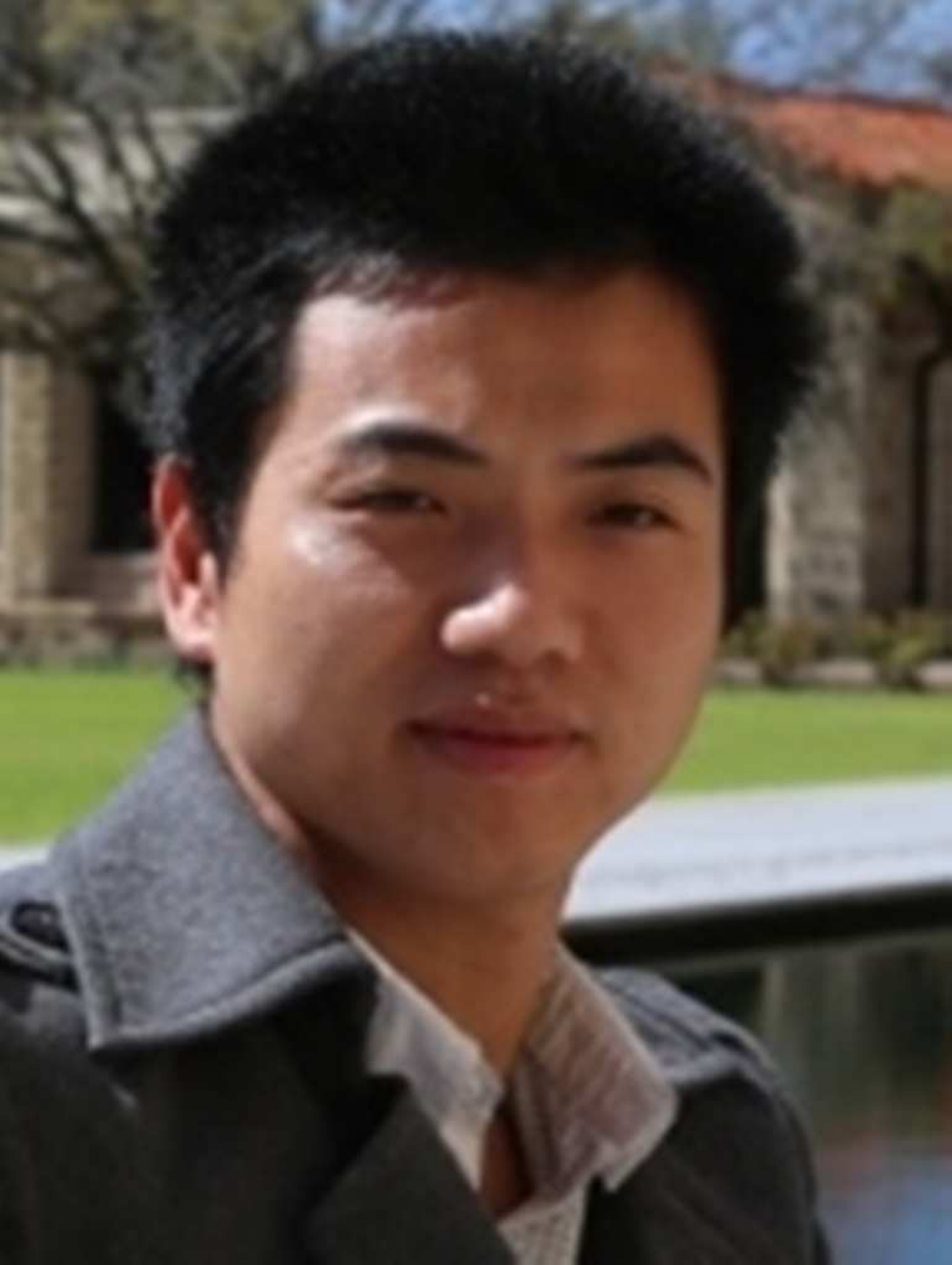}}]
	{Yulan Guo} received the B.Eng. and Ph.D. degrees from National University of Defense Technology (NUDT) in 2008 and 2015, respectively. He was a visiting Ph.D. student with the University of Western Australia from 2011 to 2014. He worked as a postdoctorial research fellow with the Institute of Computing Technology, Chinese Academy of Sciences from 2016 to 2018. He has authored over 80 articles in journals and conferences, such as the IEEE TPAMI and IJCV. His current research interests focus on 3D vision, particularly on 3D feature learning, 3D modeling, 3D object recognition, and 3D biometrics. Dr. Guo received the CAAI Outstanding Doctoral Dissertation Award in 2016. He served as an associate editor for IET Computer Vision and IET Image Processing, a guest editor for IEEE TPAMI, a PC member for several conferences (e.g., CVPR and ICCV), a reviewer for over 30 journals, and an organizer for a tutorial in CVPR 2016 and a workshop in CVPR 2019.
	\end{IEEEbiography}
	\vspace{-5ex}
	\begin{IEEEbiography}[{\includegraphics[width=1in,height=1.25in,clip]{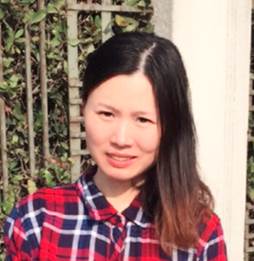}}]
		{Li Liu} received the BSc degree in communication engineering, the MSc degree in photogrammetry and remote sensing and the Ph.D. degree in information and communication engineering from the National University of Defense Technology (NUDT), Changsha, China, in 2003, 2005 and 2012, respectively. She joined the faculty at NUDT in 2012, where she is currently an Associate Professor with the College of System Engineering. During her PhD study, she spent more than two years as a Visiting Student at the University of Waterloo, Canada, from 2008 to 2010. From 2015 to 2016, she spent ten months visiting the Multimedia Laboratory at the Chinese University of Hong Kong. From 2016 to 2018, she is working at the Machine Vision Group at the University of Oulu, Finland. She was a cochair of International Workshops at ACCV2014, CVPR2016, ICCV2017 and ECCV2018. She was a guest editor of special issues for IEEE TPAMI and IJCV. Her research interests include facial behavior analysis, texture analysis, image classification, object detection and recognition.
	\end{IEEEbiography}
	\vspace{-5ex}
	\begin{IEEEbiography}[{\includegraphics[width=1in,height=1.25in,clip,keepaspectratio]{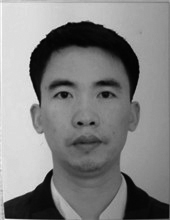}}]
		{Zaiping Lin} received the B.Eng. and Ph.D. degrees from the National University of Defense Technology (NUDT) in 2007 and 2012, respectively. He is currently an Assistant Professor with the College of Electronic Science and Technology, NUDT. His current research interests include infrared image processing and signal processing.
	\end{IEEEbiography}
	\vspace{-5ex}
	\begin{IEEEbiography}[{\includegraphics[width=1in,height=1.25in,clip,keepaspectratio]{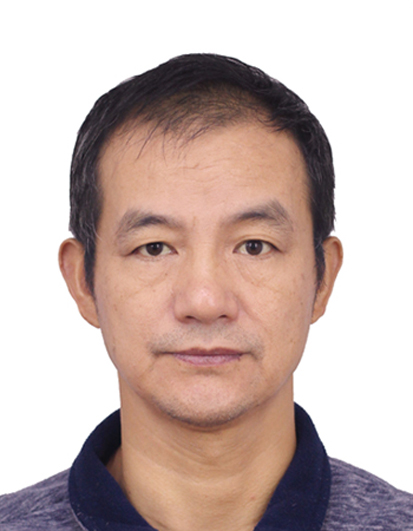}}]
		{Xinpu Deng} received the B.Eng. and Ph.D. degrees from the National University of Defense Technology (NUDT).  He is currently an Associate Professor with the College of Electronic
		Science and Technology, NUDT. His current research
		interests include signal processing and remote sensing.
	\end{IEEEbiography}
	\vspace{-5ex}
	\begin{IEEEbiography}[{\includegraphics[width=1in,height=1.25in,clip]{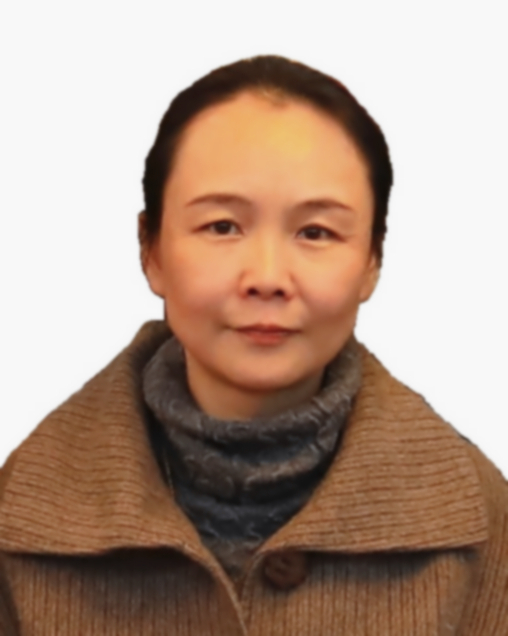}}]
		{Wei An} received the Ph.D. degree from the National University of Defense Technology (NUDT), Changsha, China, in 1999. She was a Senior Visiting Scholar with the University of Southampton, Southampton, U.K., in 2016. She is currently a Professor with the College of Electronic Science and Technology, NUDT. She has authored or co-authored over 100 journal and conference publications. Her current research interests include signal processing and image processing.
	\end{IEEEbiography}
\end{document}